\documentclass[letterpaper,twocolumn,10pt]{article}
\usepackage[top=1in, bottom=1in, left=0.9in, right=0.9in]{geometry}
\usepackage{graphicx}
\usepackage{amsmath,amssymb,amsfonts,amsthm}
\usepackage{multirow,makecell,tabularx}
\usepackage{enumitem}
\usepackage{adjustbox}
\usepackage{varwidth}
\usepackage{colortbl}

\usepackage{makecell}
\usepackage{ulem}
\usepackage{etoolbox}
\usepackage{ragged2e}
\usepackage{lmodern}
\usepackage{enumitem}

\usepackage{array}
\usepackage{multirow}
\usepackage{hhline}
\usepackage{makecell}
\usepackage{diagbox}
\usepackage{ulem}

\usepackage[table,xcdraw]{xcolor}

\newcommand{\bluebold}[1]{\textcolor{blue}{\textbf{#1}}}

\usepackage{diagbox}

\usepackage[colorlinks=true,linkcolor=blue,citecolor=blue,urlcolor=blue]{hyperref}

\title{CAMME: Adaptive Deepfake Image Detection\\ with Multi-Modal Cross-Attention}
\author{
  Naseem Khan\textsuperscript{1}, 
  Tuan Nguyen\textsuperscript{2}, 
  Amine Bermak\textsuperscript{1}, 
  Issa Khalil\textsuperscript{2} \\
  \textsuperscript{1}Hamad Bin Khalifa University (HBKU), Qatar \\
  \textsuperscript{2}Qatar Computing Research Institute (QCRI), HBKU, Qatar \\
}
\date{}

\begin{document}
\maketitle

\begin{abstract}
The proliferation of sophisticated AI-generated deepfakes poses critical challenges for digital media authentication and societal security. While existing detection methods perform well within specific generative domains, they exhibit significant performance degradation when applied to manipulations produced by unseen architectures—a fundamental limitation as generative technologies rapidly evolve. We propose CAMME (Cross-Attention Multi-Modal Embeddings), a framework that dynamically integrates visual, textual, and frequency-domain features through a multi-head cross-attention mechanism to establish robust cross-domain generalization. Extensive experiments demonstrate CAMME's superiority over state-of-the-art methods, yielding improvements of 12.56\% on natural scenes and 13.25\% on facial deepfakes. The framework demonstrates exceptional resilience, maintaining (over 91\%) accuracy under natural image perturbations and achieving 89.01\% and 96.14\% accuracy against PGD and FGSM adversarial attacks, respectively. Our findings validate that integrating complementary modalities through cross-attention enables more effective decision boundary realignment for reliable deepfake detection across heterogeneous generative architectures.

\end{abstract}

\section{Introduction}
\begin{figure}[!t]
  \centering
  \includegraphics[width=0.8\columnwidth]{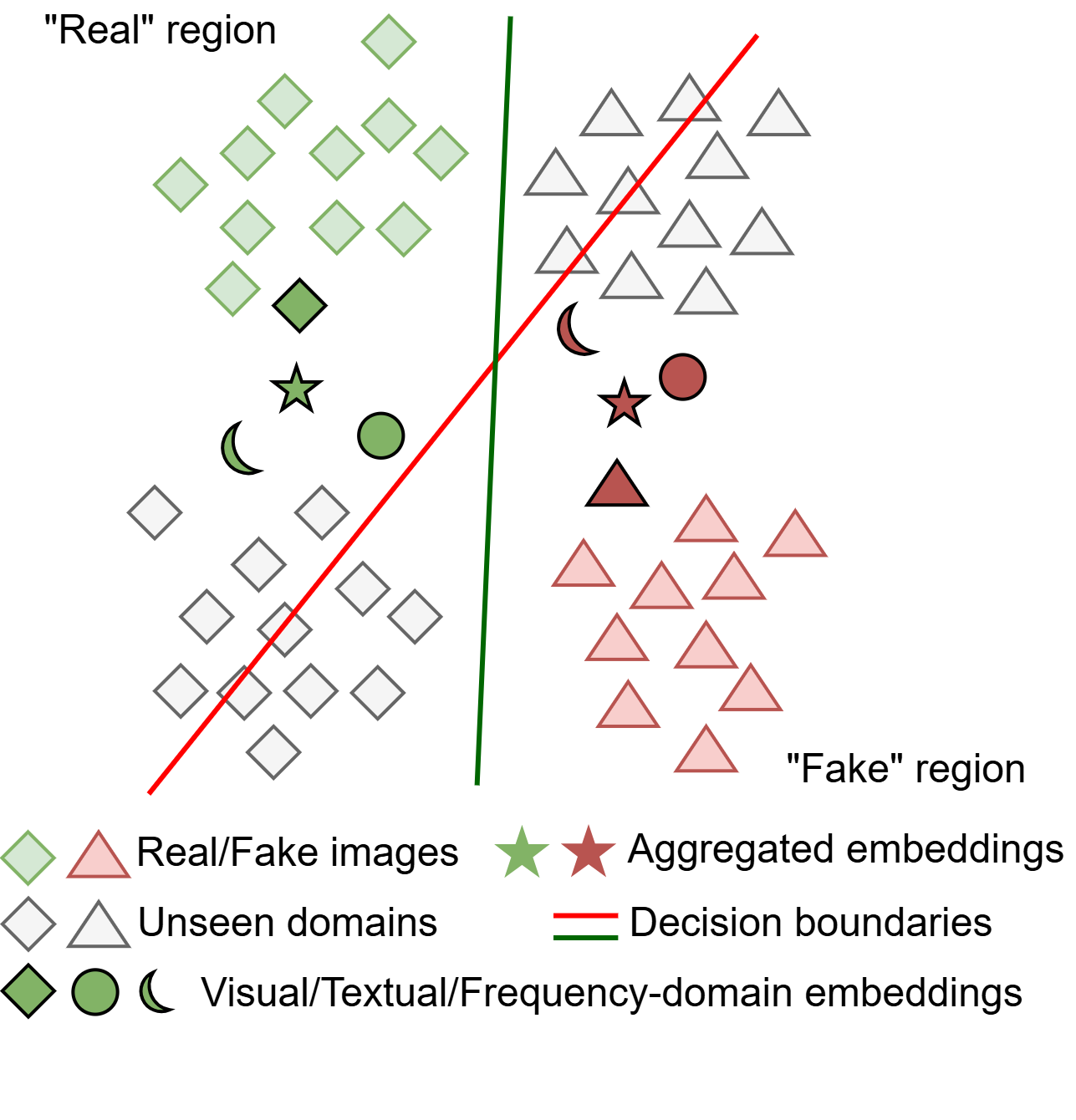} 
  \caption{Challenges in detecting unseen domains. Traditional deepfake detection methods usually rely on uni-modal embeddings (e.g., visual embeddings), resulting in non-generalizable decision boundaries (red line) that struggle with unseen domains (grey examples). Unlike previous approaches, our framework aggregates information from visual, text, and frequency domains to re-align the decision boundary (green line), enhancing transferability to unseen target domains.} 
  \label{fig:generalization} 
\end{figure}

The rise of deepfake technology—capable of synthesizing highly realistic, AI-generated static images—poses a profound challenge to digital media integrity. Leveraging Generative Artificial Intelligence (GAI), such as Stable Diffusion \cite{rombachHighResolutionImageSynthesis2022}, DALL-E \cite{rameshHierarchicalTextConditionalImage2022}, StyleCLIP \cite{patashnikStyleCLIPTextDrivenManipulation2021a}, and Imagen \cite{sahariaPhotorealisticTextImageDiffusion2022}, malicious actors can produce fabricated images that are nearly indistinguishable from real photographs. These images enable the spread of misinformation (e.g., fake news imagery), identity fraud, and unauthorized replication of artistic styles, raising ethical concerns about privacy, copyright, and originality \cite{korshunovDeepFakesNewThreat2018, mirskyCreationDetectionDeepfakes2020, nguyenDeepLearningDeepfakes2022, elgammalCANCreativeAdversarial2017, oordWaveNetGenerativeModel2016, kietzmannDeepfakesTrickTreat2020}. As GAI continues to evolve, the urgency for effective deepfake detection mechanisms grows.

Current detection methods excel in \textbf{intra-domain} settings, where training and testing occur within the same generative architecture (e.g., GAN-based models). However, real-world deepfakes often stem from diverse generative models, introducing significant distribution shifts that impair performance in \textbf{inter-domain} scenarios. As illustrated in Figure \ref{fig:generalization}, this challenge manifests in two critical ways: visually, through decision boundaries (red line) that fail to classify unseen samples correctly, and quantitatively, through severe performance degradation when models are evaluated across domains. The F1 scores presented in Table \ref{tab:test_transferability} reveal that state-of-the-art methods experience dramatic accuracy drops in cross-domain settings, with performance deteriorating from over 95\% to below 50\% when training on one generative architecture (e.g., GLIDE) and testing on another (e.g., VQDM). This evidence underscores the urgent need for detection frameworks capable of generalizing across heterogeneous generative architectures.
\begin{table}[t]
  \centering
  \caption{Cross-domain transferability analysis: F1 scores (\%) for intra-domain and inter-domain evaluations. $X \rightarrow Y$ represents training on domain $X$ and testing on domain $Y$, where G: GLIDE, V: VQDM.}
  \label{tab:test_transferability}
  \resizebox{0.95\linewidth}{!}{  
    \begin{tabular}{|c|c|c|c|c|}
      \hline
      Methods & G$\rightarrow$G & G$\rightarrow$V & V$\rightarrow$V & V$\rightarrow$G \\ \hline
      DE-FAKE \cite{shaFAKEDetectionAttribution2023} & 98.86 & 34.42 & 95.45 & 51.40 \\ \hline
      UnivCLIP \cite{ojhaUniversalFakeImage2024} & 97.63 & 28.57 & 97.39 & 43.04 \\ \hline
      UnivConv2B \cite{abdullahAnalysisRecentAdvances2024} & 99.97 & 33.17 & 99.76 & 78.70 \\ \hline
      DCT \cite{rickerDetectionDiffusionModel2024} & 99.20 & 45.84 & 97.32 & 64.45 \\ \hline
    \end{tabular}
  }
\end{table}

We propose \textbf{CAMME} (Cross-Attention Multi-Modal Embeddings), a novel framework to enhance cross-domain transferability in deepfake image detection. CAMME integrates three modalities—visual embeddings for semantic content \cite{ojhaUniversalFakeImage2024}, textual embeddings for contextual consistency \cite{shaFAKEDetectionAttribution2023}, and frequency-domain features for spectral artifacts \cite{rickerDetectionDiffusionModel2024}—to construct a robust feature space. A multi-head cross-attention mechanism, inspired by \cite{vaswaniAttentionAllYou2023}, dynamically weights these modalities, enabling adaptive focus on the most discriminative features for a given image. This approach facilitates flexible decision boundary adjustments, improving generalization to unseen domains and resilience against perturbations.

Our contributions are:
\setlength{\emergencystretch}{3em}

\begin{itemize}[leftmargin=*, labelsep=5pt, itemsep=0em]
  \RaggedRight
  \sloppy
  \item We introduce CAMME, a multi-modal cross-attention framework that enhances deepfake detection transferability across generative domains, improving state-of-the-art performance by over 12\%.
  \item We develop a multi-modal benchmark by augmenting selected images from GenImage \cite{zhuGenImageMillionScaleBenchmark2023} with BLIP-generated captions, creating a dataset of 161,837 natural scene images across five generative architectures. We also compile 100,000 facial images for comprehensive cross-domain evaluation.
  \item We demonstrate CAMME's exceptional robustness, maintaining over 91\% accuracy under natural perturbations and 89\%--96\% accuracy against adversarial attacks (PGD, FGSM).
\end{itemize}

These advancements mark a significant step toward reliable deepfake detection in an era of diverse GAI technologies. Our code and datasets are available at: \url{https://github.com/Magnet300/CAMME.git}.

\section{Related Work}
Deepfake image detection has emerged as a critical area of research due to its implications for privacy and societal security \cite{citronDeepFakesLooming2019, verdolivaMediaForensicsDeepFakes2020, masoodDeepfakesGenerationDetection2021}. We review relevant approaches and their limitations, focusing particularly on two fundamental challenges: cross-domain generalization capabilities and robustness against both natural and adversarial perturbations.

\subsection{Evolution of Deepfake Detection Methods}

Early deepfake detection relied on handcrafted feature-based methods that identified distinctive generation artifacts through preprocessing techniques \cite{xuanGeneralizationGANImage2019, wangCNNGeneratedImagesAre2020}. These approaches typically employed classical machine learning algorithms such as SVMs \cite{wangGazeLatentSupport2017}, Random Forests \cite{baiGrowingRandomForest2017}, and MLPs \cite{zhengSiameseMultilayerPerceptrons2015} for classification. However, these methods generally exhibit limited adaptability to novel generation techniques and poor cross-domain generalization \cite{masoodDeepfakesGenerationDetection2021}.

In contrast, deep feature-based methods offer more robust performance by learning representations directly from data, capturing high-level spatial patterns that adapt more effectively across different deepfake generation techniques \cite{nguyenMultitaskLearningDetecting2019, zhouFaceForensicsWild2021}. Recent state-of-the-art approaches have further enhanced detection performance by leveraging uni-modal information sources, including visual, textual, and frequency domains \cite{abdullahAnalysisRecentAdvances2024, ojhaUniversalFakeImage2024, rickerDetectionDiffusionModel2024, shaFAKEDetectionAttribution2023}.

\subsection{Uni-Modal Approaches and Their Limitations}

Recent deepfake detection methods can be categorized based on their primary feature domain. Visual embedding-based approaches like UnivCLIP \cite{ojhaUniversalFakeImage2024} leverage CLIP \cite{radfordLearningTransferableVisual2021} to extract semantic representations, measuring distances between image embeddings for classification. While effective within domains, these approaches struggle with cross-domain generalization due to their limited ability to identify subtle artifacts across different generative architectures.

Text-augmented approaches such as DE-FAKE \cite{shaFAKEDetectionAttribution2023} combine visual and textual embeddings using CLIP and caption generation models like BLIP \cite{liBLIPBootstrappingLanguageImage2022}. These methods exploit the observation that real images typically contain more diverse and informative content than synthetic images generated from simpler prompts. However, they remain vulnerable to sophisticated generations that closely match prompt descriptions.

Frequency-domain methods like DCT \cite{rickerDetectionDiffusionModel2024} focus on spectral artifacts, especially useful for detecting diffusion model outputs that exhibit fewer spatial anomalies than GAN-generated images. By applying logarithmic Discrete Cosine Transform analysis, these approaches have demonstrated significant improvements (16.3\%) over visual-only methods. Nevertheless, they often lack robustness against adversarial attacks that specifically target frequency patterns.

Comprehensive approaches like UnivConv2B \cite{abdullahAnalysisRecentAdvances2024} have analyzed trade-offs between generalization performance and adversarial robustness. Their findings reveal that frequency-domain models excel in generalization but remain vulnerable to adversarial manipulation, while visual feature-based models show greater resilience against attacks at the cost of reduced cross-domain transferability.

\subsection{Watermarking vs. Artifact-Based Detection}

An alternative to artifact detection is watermarking \cite{wangAvisual2015, HasanImage2021, yingFrom2021, yingLearning2023, anWAVES2024}, which embeds imperceptible patterns during image creation for later verification. However, watermarking approaches face significant limitations: vulnerability to common image processing operations (resizing, transcoding, filtering), susceptibility to manipulation by advanced AI tools, image quality degradation, and challenges in balancing robustness with imperceptibility.

\subsection{Motivation for Multi-Modal Approach}

Our analysis of existing uni-modal approaches reveals a critical gap: while each domain (visual, textual, frequency) provides valuable detection signals, none alone achieves both strong cross-domain generalization and robust adversarial defense. As demonstrated in Figure \ref{fig:generalization}, state-of-the-art methods exhibit severe performance collapse ($>60\%$ F1-score drop) when evaluated across different generative domains. This limitation motivates our multi-modal framework, which integrates complementary features through cross-attention mechanisms to dynamically adjust decision boundaries, addressing fundamental generalization challenges in deepfake detection.

\section{Background}
\subsection{Generative Models}
Generative Adversarial Networks (GANs) \cite{goodfellowGenerativeAdversarialNetworks2014,choiStarGANUnifiedGenerative2018,karrasProgressiveGrowingGANs2018,brockLargeScaleGAN2019} were initially developed to synthesize images by learning a distribution that closely mimics that of real data. GANs have since been widely adopted across various applications, particularly for image generation. However, GAN-based models often face challenges in producing realistic images due to issues like training instability and mode collapse, where the generator fails to capture the full diversity of the data. These limitations have led to the emergence of diffusion models as a preferred alternative, offering improved realism in generated images. Denoising Diffusion Probabilistic Models (DDPMs) \cite{hoDenoisingDiffusionProbabilistic2020} pioneered this approach, successfully generating high-quality images through a denoising diffusion process and inspiring a wide range of diffusion models \cite{rombachHighResolutionImageSynthesis2022,dhariwalDiffusionModelsBeat2021,nicholGLIDEPhotorealisticImage2022,guVectorQuantizedDiffusion2022}.

Recent advancements, such as Stable Diffusion, introduced cross-attention mechanisms between text and visual embeddings, highlighting the effectiveness of text-driven image generation by focusing on user prompts. This integration of text and image information has been further refined in the CLIP model \cite{radfordLearningTransferableVisual2021}, which employs a contrastive loss to align text and image embeddings. By anchoring semantic meaning to images, CLIP not only enhances zero-shot classification performance but also strengthens the connection between textual and visual modalities for diverse downstream tasks.

\subsection{Transfer Learning}
In this section, we introduce the framework of transfer learning for deepfake image detection. In transfer learning or domain adaptation, the objective is to adapt a model trained on one domain (referred to as the source domain) so that it performs well on a different domain (referred to as the target domain). For instance, in the context of deepfake image detection, the source domain could be a set of images generated by a GAN-based model, while the target domain could consist of images generated by an unseen diffusion-based model. 

Let $\mathcal{D}^S = \{(x_i^S, y_i^S)\}_{i=1}^{N_S}$ represent the dataset of the source domain, where $x_i^S \in \mathcal{X}^S$ denotes the input features and $y_i^S \in \mathcal{Y}$ denotes the labels. In deep fake detection, $y_i = 0$ represents the \textit{real} class, and $y_i = 1$ represents the \textit{fake} class. The dataset for the target domain is denoted by $\mathcal{D}^T = \{x_i^T\}_{i=1}^{N_T}$. Note that the target domain remains entirely unseen and unlabeled during training.

Our primary objective is to learn a mapping \( f: \mathcal{X} \rightarrow \mathcal{Y} \) that generalizes well on the source domain and performs robustly on the unseen target domain:

\begin{equation}
\min_{f} \mathbb{E}_{(x, y) \sim P^T} \left[ \mathcal{L}(f(x), y) \right],
\end{equation}

where \( P^T \) represents the distribution over the target domain, and \( \mathcal{L} \) is the loss function (e.g., cross-entropy loss). However, in this context, we do not have access to labeled target data during training, so we cannot directly minimize the above objective. Instead, we approximate the mapping \( f \) by minimizing the loss over the source domain distribution \( P^S \), where labeled data is available:

\begin{equation} \label{eq:source_loss}
\min_{f} \mathbb{E}_{(x, y) \sim P^S} \left[ \mathcal{L}(f(x), y) \right].
\end{equation}

In practice, \( f \) can be decomposed as \( f = h \circ G \), where \( G: \mathcal{X} \rightarrow \mathcal{Z} \) maps the input \( x \) to a feature space \( \mathcal{Z} \) (referred to as the feature extractor), and \( h: \mathcal{Z} \rightarrow \mathcal{Y} \) maps from the feature space to the final output in \( \mathcal{Y} \) (referred to as the classifier). The objective in Equation (\ref{eq:source_loss}) can then be rewritten as: 

\begin{equation} \label{eq:source_loss_decomposed}
\min_{G, h} \mathbb{E}_{(x, y) \sim P^S} \left[ \mathcal{L}(h(G(x)), y) \right].
\end{equation}

\subsection{Discrete Cosine Transform}
The Discrete Cosine Transform (DCT) is widely used in image processing and computer vision tasks due to its ability to capture frequency components effectively \cite{ahmedDiscreteCosineTransform1974, raoDiscreteCosineTransform1990, wallaceJPEGSTILLPICTURE}. An image \( I \) can be transformed into its DCT representation \( F \) using the following equation:

\begin{equation} \label{eq:dct}
    F_{k,l} = \sum_{m=0}^{M-1} \sum_{n=0}^{N-1} I_{m,n} \, C_M(m, k) \, C_N(n, l),
\end{equation}

where
\begin{align}
&C_M(m, k) = \cos \left( \frac{\pi}{M} \left( m + \frac{1}{2} \right) k \right), \\
&C_N(n, l) = \cos \left( \frac{\pi}{N} \left( n + \frac{1}{2} \right) l \right).
\end{align}

Here, \( F_{k,l} \) represents the DCT coefficient at row \( k \) and column \( l \) of the DCT matrix, while \( I_{m,n} \) denotes the pixel value at row \( m \) and column \( n \) in the image. The DCT coefficients capture various frequency components of the image, with \textit{lower frequencies} concentrated in the top-left corner of the DCT matrix and \textit{higher frequencies} towards the bottom-right. Lower frequencies correspond to smooth, gradual intensity changes (e.g., background regions), whereas higher frequencies capture rapid changes (e.g., edges and textures). This property makes DCT particularly useful for detecting artifacts in image forensics and deepfake detection, as real and manipulated images often exhibit distinct patterns in their frequency components.

\begin{figure*}[!t]
    \centering
    \makebox[\textwidth][c]{
        \includegraphics[width=\textwidth]{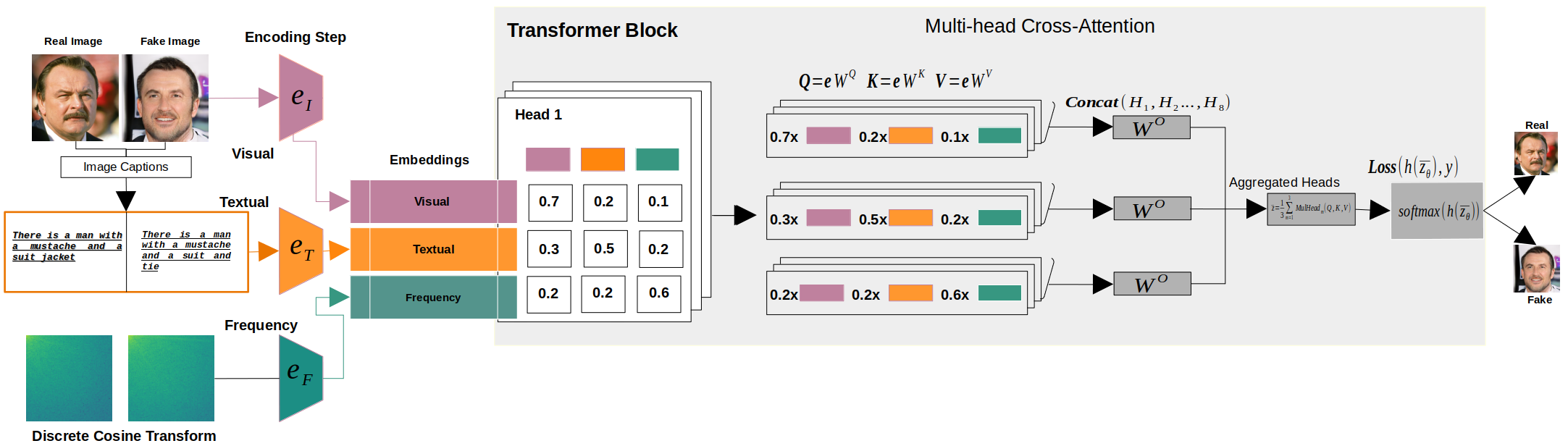}}
        \caption{CAMME framework: A multi-modal transformer architecture for deepfake detection. Images are encoded into visual, textual, and frequency domain embeddings, which are treated as sequential tokens. The transformer block applies self-attention across these tokens, enabling cross-modal interactions. Each attention head dynamically weights features from different modalities. The final representation is obtained by weighted aggregation of the attention outputs for binary classification. The model effectively captures complementary discriminative features across domains.}
       \label{fig:framework}
\end{figure*}

\section{Our Proposed Method}
\subsection{Overview}
In this section, we introduce our proposed method to enhance transferability in deepfake image detection. Current state-of-the-art detectors primarily rely on uni-modal embeddings, such as \textit{image-based} \cite{ojhaUniversalFakeImage2024,abdullahAnalysisRecentAdvances2024} or \textit{frequency-based} modalities \cite{rickerDetectionDiffusionModel2024}, each offering distinct advantages. For instance, an analysis in \cite{abdullahAnalysisRecentAdvances2024} indicates that frequency-domain-based models exhibit strong generalization, whereas models based on visual features demonstrate greater resilience to adversarial manipulations. The main challenge lies in integrating these uni-modal embeddings to enhance generalization and transferability across diverse domains, while preserving the unique benefits of each embedding type. Furthermore, connecting embeddings from different spaces is complex, as each type (e.g., visual, textual, and frequency-domain) has distinct characteristics and representations.
Motivated by this challenge, we propose a multi-modal deepfake detection method that integrates visual, textual, and frequency-domain features, leveraging the strengths of these uni-modal embeddings through a cross-attention mechanism. Our approach, named \textbf{CAMME} (\textbf{C}ross-\textbf{A}ttention on \textbf{M}ulti-\textbf{M}odal \textbf{E}mbedding), is designed to enhance deepfake detection capabilities and improve model transferability across diverse domains.

\subsection{Modality Embedding Extraction}
In what follows, we describe the feature extraction process for visual, text, and frequency-domain embeddings.

\textbf{Visual Embeddings.} Rather than designing vision models from scratch, we leverage pre-trained visual models trained on large-scale datasets. These models learn highly generalizable representations, producing rich features suitable for downstream tasks such as image classification. For our visual feature extraction, we adopt OpenCLIP-ConvNextLarge \cite{LaionCLIPconvnext_large_d_320laion2Bs29Bb131KftsoupHugging2023}, following established baselines \cite{abdullahAnalysisRecentAdvances2024} for feature extractors.

\textbf{Text Embeddings.} Real images typically contain richer semantic content than the prompts used to generate fake images, which are often derived from simpler prompts. Based on this observation, \cite{shaFAKEDetectionAttribution2023} demonstrates that textual information can help distinguish between real and fake images. In this work, we utilize CLIP’s text encoder to extract embeddings from image prompts. For datasets lacking prompt information, we employ BLIP \cite{liBLIPBootstrappingLanguageImage2022} to generate captions for the images, making this approach applicable to real-world scenarios where associated text may be unavailable.

\textbf{Frequency-Domain Embeddings.} Detecting fake images in the frequency domain has shown potential, as generated images often display differences (e.g., artifacts) from real images. Inspired by \cite{rickerDetectionDiffusionModel2024}, we apply the Discrete Cosine Transform (DCT) to extract frequency-domain features from images, as described in Equation (\ref{eq:dct}). These frequency-based features capture artifacts that may be indicative of image manipulation.

\subsection{Multi-head Cross Attention Mechanism}
To find the optimal solution for Equation (\ref{eq:source_loss_decomposed}), we first extract visual, textual, and frequency-domain embeddings from the input images:

\begin{equation}
    \mathcal{\textbf{e}}_I = G_{I}(x), \quad \mathcal{\textbf{e}}_T = G_{T}(x), \quad \mathcal{\textbf{e}}_F = G_{F}(x),
\end{equation}

where \( G_{I}, G_{T}, G_{F} \) denote the feature extractors for visual, textual, and frequency-domain embeddings, respectively. These embeddings raise the critical question of how to effectively integrate them to enhance model generalization. Various techniques, such as element-wise addition, concatenation, average pooling, or learnable weighted averaging, can be applied to combine these embeddings. However, as shown in our experiments (Section \ref{sec:embedding_combination}), these methods may not fully capture the interactions among embeddings. Additionally, the model's performance may depend on the type of information needed for each detection. For instance, ``easy" fake images, such as those generated by GAN-based models, are often distinguishable using frequency-domain features alone \cite{rickerDetectionDiffusionModel2024}, whereas ``harder" examples may require multi-modal information for accurate detection. Therefore, the model's detection performance is highly dependent on its ability to selectively focus on relevant information from each embedding type.

Inspired by this insight, we propose a cross-attention mechanism to strengthen the interactions among uni-modal embeddings and enable the model to focus on the most relevant information for each decision. Under the framework of the multi-head self-attention architecture \cite{vaswaniAttentionAllYou2023}, we treat each uni-modal embedding as a \textit{token} in a sequence to apply attention across these uni-modal tokens. This viewpoint maintains flexibility, enabling our model to be extended to more than three modalities, such as visual, text, frequency, audio, sensor, or graph data, by treating each modality as a distinct token in the sequence. In this framework, let \( \textbf{e} \) represent a sequence of three token embeddings:

\begin{equation} \label{eq:token_embedding}
    \textbf{e} = \{\textbf{e}_I, \textbf{e}_T, \textbf{e}_F\} \in \mathbb{R}^{3 \times d_e}.
\end{equation}

We then project \( \textbf{e} \) into query (\( Q \)), key (\( K \)), and value (\( V \)) matrices as follows:

\begin{equation}
    Q = \textbf{e}W^Q, \quad K = \textbf{e}W^K, \quad V = \textbf{e}W^V,
\end{equation}

where \( Q \in \mathbb{R}^{3 \times d_k} \), \( K \in \mathbb{R}^{3 \times d_k} \), and \( V \in \mathbb{R}^{3 \times d_v} \).

The self-attention mechanism \cite{vaswaniAttentionAllYou2023} is applied to \( Q \), \( K \), and \( V \), with the attention score computed as:

\begin{equation}
    \text{Attention}(Q, K, V) = \text{softmax}\left(\frac{QK^\top}{\sqrt{d_k}}\right)V.
\end{equation}

Here, \( QK^\top \) measures the similarity between token embeddings in \( \textbf{e} \). Thus, \textit{the self-attention mechanism applied to \( \textbf{e} \) acts as a form of cross-attention across the visual, text, and frequency-domain embeddings}. In multi-head attention, we concatenate the outputs of multiple attention heads and project them back to the original embedding dimension. Let \( h_i = \text{head}_i \) and \( \text{MH} = \text{MultiHead} \) for brevity:
\begin{align}
    h_i &= \text{Attention}(Q_i, K_i, V_i), \\
    \text{MH}(Q, K, V) &= \text{Concat}(h_1, \ldots, h_H)W^O
\end{align}
where \( h_i \in \mathbb{R}^{3 \times d_v} \), \( W^O \in \mathbb{R}^{Hd_v \times d_e} \) is the output projection matrix, and \( H \) is the number of heads. This yields \( \text{MH}(Q, K, V) \in \mathbb{R}^{3 \times d_e} \). The output of the multi-head attention is then aggregated into a single embedding that captures the interactions among the visual, text, and frequency-domain embeddings:
\begin{equation}
    \bar{z} = \frac{1}{3} \sum_{m=1}^3 \text{MH}_m(Q, K, V).   
\end{equation}

\subsection{Training Objective}

The aggregated embedding \( \bar{z} \) is subsequently fed into a classifier $h$ to predict the probability that the input image is real or fake. The optimization problem in Equation (\ref{eq:source_loss_decomposed}) is then reformulated as:

\begin{equation} \label{eq:source_loss_final}
    \min_{\theta, h} \mathbb{E}_{(x, y) \sim P^S} \left[ \mathcal{L}(h(\bar{z}_\theta), y) \right],
\end{equation}

where \( \theta = \{W^Q_i, W^K_i, W^V_i, W^O\}\) are the learnable parameters in the multi-head attention module.  Details of the training process are provided in Section \ref{sec:implementation}. Figure \ref{fig:framework} illustrates our proposed framework for deepfake image detection.

\section{Experiment}
In this section, we present our data collection and experimental results on natural scene and face datasets.

\subsection{Dataset Construction} \label{sec:datasets}
\subsubsection{Natural Scene Dataset}
To evaluate the transferability of our model, we follow the approach in \cite{zhuGenImageMillionScaleBenchmark2023} to construct a dataset using images generated by popular generative models, including Stable Diffusion V1.5 \cite{rombachHighResolutionImageSynthesis2022}, ADM \cite{dhariwalDiffusionModelsBeat2021}, GLIDE \cite{nicholGLIDEPhotorealisticImage2022}, VQDM \cite{guVectorQuantizedDiffusion2022}, and BigGAN \cite{brockLargeScaleGAN2019}. The authors generated synthetic images using pre-trained models with prompts in the format ``photo of [class name]", where ``[class name]" is replaced with ImageNet labels.  We collect in total 161,837 images from these models.

Following the procedure outlined in \cite{abdullahAnalysisRecentAdvances2024}, we split each dataset into training, validation, and test sets with an \(80:10:10\) ratio. The dataset is balanced, containing an equal number of real and fake images. Each dataset includes 100 categories, specifically the first 100 categories from ImageNet \cite{dengImageNetLargeScaleHierarchical2009}. With around 320 images per category, this dataset poses a significant challenge for any deepfake detection method to generalize effectively across unseen domains. 

\subsubsection{Face Dataset}
The face dataset is based on the dataset introduced in \cite{guoFakeFaceDetection2020}. Real images are sourced from CelebA \cite{liuDeepLearningFace}, CelebA-HQ \cite{karrasProgressiveGrowingGANs2018}, and Youtube-Frame \cite{rosslerFaceForensicsLearningDetect2019}, while fake images are generated using PGGAN \cite{karrasProgressiveGrowingGANs2018}, Glow \cite{kingmaGlowGenerativeFlow2018}, Face2Face \cite{thiesFace2FaceRealtimeFace2020}, and StarGAN \cite{choiStarGANUnifiedGenerative2018}. To represent the face domain comprehensively, we create the following combined subsets: (i) CelebA-HQ paired with PGGAN (CelPG); (ii) CelebA with StarGAN (CelStar); (iii) CelebA with Glow (CelGlow); and (iv) Youtube-Frame with Face2Face (YouFace). These combinations align the properties of real and synthetic images within each subset, enabling meaningful comparisons between real and fake classes. For each dataset, we collect 20,000 images, consisting of 10,000 real and 10,000 fake images, resulting in a total of 100,000 images for the face dataset.

The natural scene and face datasets encompass a diverse range of generative models and content types, making them representative of real-world deepfake challenges. The Natural Scene Dataset captures variability across generative architectures, with models producing highly realistic images that are challenging to detect due to subtle artifacts. The Face Dataset includes both unaltered and manipulated faces, representing common deepfake targets such as identity swapping and attribute editing. These datasets were selected to address key challenges: generalization to unseen deepfakes, robustness against high-quality manipulations, and adaptation to diverse content types. Together, they provide a comprehensive benchmark for advancing deepfake detection methods and tackling future challenges posed by increasingly sophisticated deepfakes. The two datasets collectively comprise 261,837 images. Appendix \ref{app:dataset_details} provides further details on the composition and distribution of the datasets, along with illustrative examples of real and fake images from each dataset (Figures \ref{fig:combined_real_and_fake}).

\begin{table*}[t]
    \centering 
    \caption{F1 score (\%) for detection performance on the natural scene dataset. Bold blue text indicates intra-domain performance. The highest score is shown in bold black, and the second-highest is underlined.}
    \label{table:natural_scene_results}
    \renewcommand{\arraystretch}{1.3}
    \resizebox{\textwidth}{!}{%
    \begin{tabular}{|c|c|c|c|c|c|c|c|c|}
    \hline
    \textbf{Methods}  & \textbf{\backslashbox[20mm]{Source}{Target}}  & \textbf{SD V1.5}  & \textbf{ADM}  & \textbf{GLIDE}  & \textbf{VQDM} & \textbf{BigGAN}  & \textbf{\makecell{Inter-domain\\Average (IA)} } & \textbf{Average of IA}\tabularnewline
    \hline
    \multirow{5}{*}{DE-FAKE}  & SD V1.5  & \textbf{\textcolor{blue}{98.62 }} & 1.96  & 52.34  & 22.06 & 7.39  & 20.94 & \multirow{5}{*}{33.29}\tabularnewline
    \cline{2-8}
     & ADM  & 1.86  & \textbf{\textcolor{blue}{98.53 }} & 13.83  & 04.53 & 27.92  & 12.04 & \tabularnewline
    \cline{2-8}
     & GLIDE  & 42.89  & 28.47  & \textbf{\textcolor{blue}{98.86 }} & 34.42 & 94.74  & 50.13 & \tabularnewline
    \cline{2-8}
     & VQDM  & 34.64  & 23.16  & 51.40  & \textbf{\textcolor{blue}{95.45}} & 78.32  & 46.88 & \tabularnewline
    \cline{2-8}
     & BigGAN  & 4.05  & 26.71  & 93.70  & 21.31 & \textbf{\textcolor{blue}{99.62 }} & 36.44 & \tabularnewline
    \hline
    \multirow{5}{*}{UnivCLIP}  & SD V1.5  & \textbf{\textcolor{blue}{97.41 }} & 7.09  & 65.55  & 14.60 & 34.93  & 30.54 & \multirow{5}{*}{41.53}\tabularnewline
    \cline{2-8}
     & ADM  & 10.27  & \textbf{\textcolor{blue}{92.76 }} & 71.13  & 93.89 & 83.76  & 64.76 & \tabularnewline
    \cline{2-8}
     & GLIDE  & 41.45  & 27.15  & \textbf{\textcolor{blue}{97.63 }} & 28.57 & 56.93  & 38.53 & \tabularnewline
    \cline{2-8}
     & VQDM  & 16.96  & 53.79  & 43.04  & \textbf{\textcolor{blue}{97.39}} & 88.33  & 50.53 & \tabularnewline
    \cline{2-8}
     & BigGAN  & 2.54  & 9.55  & 22.96  & 58.02 & \textbf{\textcolor{blue}{99.09}} & 23.27 & \tabularnewline
    \hline
    \multirow{5}{*}{UnivConv2B}  & SD V1.5  & \textbf{\textcolor{blue}{99.86 }} & 0.12  & 12.86  & 01.52 & 4.60  & 4.78 & \multirow{5}{*}{39.71}\tabularnewline
    \cline{2-8}
     & ADM  & 0.78  & \textbf{\textcolor{blue}{99.71 }} & 99.12  & 98.91 & 99.74  & \uline{74.64} & \tabularnewline
    \cline{2-8}
     & GLIDE  & 1.12  & 24.38  & \textbf{\textcolor{blue}{99.97 }} & 33.17 & 70.87  & 32.39 & \tabularnewline
    \cline{2-8}
     & VQDM  & 4.49  & 90.28  & 78.70  & \textbf{\textcolor{blue}{99.76}} & 95.57  & 67.26 & \tabularnewline
    \cline{2-8}
     & BigGAN  & 2.66  & 10.52  & 24.65  & 40.06 & \textbf{\textcolor{blue}{99.94 }} & 19.47 & \tabularnewline
    \hline
    \multirow{5}{*}{DCT}  & SD V1.5  & \textbf{\textcolor{blue}{96.74 }} & 46.52  & 63.11  & 90.27 & 52.86  & \uline{63.19} & \multirow{5}{*}{\uline{65.15}}\tabularnewline
    \cline{2-8}
     & ADM  & 72.17  & \textbf{\textcolor{blue}{99.20 }} & 63.42  & 72.64 & 51.27  & 64.88 & \tabularnewline
    \cline{2-8}
     & GLIDE  & 50.68  & 51.07  & \textbf{\textcolor{blue}{99.20 }} & 45.84 & 72.19  & \uline{54.95} & \tabularnewline
    \cline{2-8}
     & VQDM  & 86.01  & 73.71  & 64.45  & \textbf{\textcolor{blue}{97.32}} & 72.74  & \uline{74.23} & \tabularnewline
    \cline{2-8}
     & BigGAN  & 57.04  & 63.06  & 96.74  & 57.19 & \textbf{\textcolor{blue}{100.00 }} & 68.51 & \tabularnewline
    \hline
    \multirow{5}{*}{CAMME (Ours)}  & SD V1.5  & \textbf{\textcolor{blue}{99.49 }} & 66.47  & 74.76  & 68.35 & 72.03  & \textbf{70.40 (+7.21)} & \multirow{5}{*}{\textbf{77.71 (+12.56)}}\tabularnewline
    \cline{2-8}
     & ADM  & 66.19  & \textbf{\textcolor{blue}{99.47 }} & 95.65  & 98.97 & 99.26  & \textbf{90.02 (+15.38)} & \tabularnewline
    \cline{2-8}
     & GLIDE  & 66.94  & 69.35  & \textbf{\textcolor{blue}{99.65 }} & 73.41 & 74.12  & \textbf{70.96 (+16.01)} & \tabularnewline
    \cline{2-8}
     & VQDM  & 66.73  & 94.12  & 84.60  & \textbf{\textcolor{blue}{99.79}} & 92.39  & \textbf{84.46 (+10.23)} & \tabularnewline
    \cline{2-8}
     & BigGAN  & 67.08  & 70.48  & 74.94  & 78.40 & \textbf{\textcolor{blue}{99.76 }} & \textbf{72.73 (+4.22)} & \tabularnewline
    \hline
    \end{tabular}
    }
\end{table*}

\subsection{Implementation Details} \label{sec:implementation}
\subsubsection{Training}
Images are preprocessed and resized to $320\times320$ to match the input requirements of the pretrained OpenCLIP-ConvNextLarge model \cite{radfordLearningTransferableVisual2021}. Text prompts are also processed through the same pretrained model to obtain text embeddings. Frequency features are extracted by applying the Discrete Cosine Transform (DCT) to the images, following \cite{abdullahAnalysisRecentAdvances2024,rickerDetectionDiffusionModel2024}, where we compute the logarithm of the absolute values of the DCT coefficients to capture frequency-domain characteristics. The DCT features are then normalized by subtracting the mean and dividing by the standard deviation, followed by a linear transformation to match the dimension of the visual and text embeddings. 

At the end of the feature extraction process, we obtain three embeddings—visual, text, and frequency-domain—with a unified dimension of $d_e = 768$. These uni-modal embeddings are subsequently fed into the multi-head cross-attention module with 8 heads, where the dimensions of $Q$, $K$, and $V$ are set to $d_k = d_v = 96$. The model computes attention scores using scaled dot-product attention to capture cross-modal relationships. After applying softmax to obtain attention weights, these weights are used to compute a weighted average of the value \( V \), effectively integrating information across domains. The outputs from each head are concatenated and averaged to produce the aggregated embedding, which is expected to capture diverse interactions among visual, text, and frequency-domain features. Finally, this aggregated embedding is passed through a classifier to predict whether the input image is real or fake. The model is trained by minimizing the loss in Equation (\ref{eq:source_loss_final}) using backpropagation. For fair comparison with baseline implementations in \cite{abdullahAnalysisRecentAdvances2024,ojhaUniversalFakeImage2024}, we freeze the pretrained visual and text embeddings from OpenCLIP-ConvNextLarge during training.

For the baseline models, we use the authors' official code repositories and evaluate these methods on our dataset with the same hyperparameters specified in their original implementations. For our proposed method, we use a batch size of 128 and the Adam optimizer \cite{kingmaAdamMethodStochastic2017} with a learning rate of \(0.001\). Training is conducted for 30 epochs with early stopping based on validation loss. All experiments are conducted on 8 Tesla V100-SXM2 GPUs with 32GB of memory each.

\subsubsection{Evaluation} The evaluation is performed on the test set, with Precision, Recall, F1 score, and Accuracy reported as evaluation metrics. For consistency, F1 scores for the Natural Scene and Face datasets are provided in Table \ref{table:natural_scene_results} and Table \ref{table:face_results_simple}, respectively, as F1 offers a balanced metric between Precision and Recall. Detailed results for Precision, Recall, F1, and Accuracy are presented in Appendix \ref{app:full_results}.

We also report intra-domain performance in the tables, representing the model's performance within the same domain. Additionally, the Inter-domain Average (IA) is calculated as the average of all inter-domain transfer tasks (where the source and target domains differ) to measure model performance on unseen domains. The overall transferability performance of the model on unseen datasets is represented by the Average of IA.
\begin{table*}[!t]
\centering
\caption{F1 score (\%) for detection performance on the face dataset (Simplified). Bold
blue text indicates intra-domain performance. The highest score is
shown in bold black, and the second-highest is underlined.}
\label{table:face_results_simple}

\renewcommand{\arraystretch}{1.3}
\resizebox{\textwidth}{!}{%
\begin{tabular}{|c|c|c|c|c|c|c|c|}
\hline
\textbf{Methods}  & \diagbox{\textbf{Source}}{\textbf{Target}}  & \textbf{CelPG}  & \textbf{CelStar}  & \textbf{CelGlow}  & \textbf{YouFace}  & \textbf{\makecell{Inter-domain\\Average (IA)}} & \textbf{Average of IA} \\
\hline
\multirow{4}{*}{DE-FAKE} & CelPG  & \textbf{\textcolor{blue}{95.91 }} & 63.30  & 7.03  & 1.59  & 23.97 & \multirow{4}{*}{15.98} \\
\cline{2-7}
 & CelStar  & 2.37  & \textbf{\textcolor{blue}{99.85 }} & 5.63  & 0.00  & 2.67 & \\
\cline{2-7}
 & CelGlow  & 0.00  & 0.40  & \textbf{\textcolor{blue}{99.24 }} & 1.09  & 0.5 & \\
\cline{2-7}
& YouFace  & 67.00  & 39.49  & 3.89  & \textbf{\textcolor{blue}{80.72 }} & 36.79 & \\
\hline
\multirow{4}{*}{UnivCLIP} & CelPG  & \textbf{\textcolor{blue}{99.60 }} & 92.28  & 31.74  & 58.21  & \uline{60.74} & \multirow{4}{*}{\uline{53.21}} \\
\cline{2-7}
 & CelStar  & 77.03  & \textbf{\textcolor{blue}{99.25 }} & 6.00  & 40.73  & 41.25 & \\
\cline{2-7}
 & CelGlow  & 6.73  & 72.03  & \textbf{\textcolor{blue}{98.08 }} & 31.55  & 36.77 & \\
\cline{2-7}
& YouFace  & 93.31  & 87.44  & 41.47  & \textbf{\textcolor{blue}{92.81 }} & \textbf{74.07} & \\
\hline
\multirow{4}{*}{UnivConv2B} & CelPG  & \textbf{\textcolor{blue}{99.90 }} & 78.80  & 14.14  & 30.55  & 41.16 & \multirow{4}{*}{28.94} \\
\cline{2-7}
 & CelStar  & 0.00  & \textbf{\textcolor{blue}{99.95 }} & 0.00  & 0.00  & 0.00 & \\
\cline{2-7}
 & CelGlow  & 0.79  & 67.29  & \textbf{\textcolor{blue}{100.00 }} & 34.66  & 34.25 & \\
\cline{2-7}
& YouFace  & 79.90  & 13.81  & 27.35  & \textbf{\textcolor{blue}{98.27 }} & 40.35 & \\
\hline
\multirow{4}{*}{DCT} & CelPG  & \textbf{\textcolor{blue}{92.42 }} & 9.14  & 41.49  & 38.04  & 29.56 & \multirow{4}{*}{44.65} \\
\cline{2-7}
 & CelStar  & 34.85  & \textbf{\textcolor{blue}{99.95 }} & 54.26  & 57.93  & \uline{49.01} & \\
\cline{2-7}
 & CelGlow  & 44.59  & 20.21  & \textbf{\textcolor{blue}{99.35 }} & 53.43  & \uline{39.41} & \\
\cline{2-7}
& YouFace  & 55.68  & 76.10  & 50.05  & \textbf{\textcolor{blue}{75.72 }} & 60.61 & \\
\hline
\multirow{4}{*}{CAMME (Ours)} & CelPG  & \textbf{\textcolor{blue}{99.75 }} & 75.17  & 62.24  & 70.67  & \textbf{69.36 (+8.62)} & \multirow{4}{*}{\textbf{66.46 (+13.25)}} \\
\cline{2-7}
 & CelStar  & 66.69  & \textbf{\textcolor{blue}{100.00 }} & 66.73  & 66.60  & \textbf{66.67 (+17.66)} & \\
\cline{2-7}
 & CelGlow  & 66.78  & 77.35  & \textbf{\textcolor{blue}{99.95 }} & 55.74  & \textbf{66.62 (+27.21)} & \\
\cline{2-7}
& YouFace  & 69.34  & 54.68  & 55.40  & \textbf{\textcolor{blue}{95.32 }} & \uline{63.19} & \\
\hline
\end{tabular}
}
\end{table*}
\subsection{Results and Discussion}
\subsubsection{Natural Scene Dataset}
Baseline methods such as DE-FAKE, UnivConv2B, and DCT perform well in source-only settings, where the model is trained and evaluated on the same dataset (e.g., SD V1.5 $\rightarrow$ SD V1.5, ADM $\rightarrow$ ADM). DCT, in particular, achieves strong results, correctly identifying BigGAN images with 100\% accuracy. However, these methods exhibit a substantial drop in average F1 scores when applied to unseen target domains. For instance, when detecting fake images from an unseen model after training on SD V1.5, DE-FAKE averages an F1 score of 20.94\%, while UnivCLIP achieves 30.54\%.

In contrast, our proposed CAMME achieves the highest inter-domain average F1 score across all source-target combinations, with improvements of 7.21\% when trained on SD V1.5, 15.38\% when trained on ADM, and a significant increase of 16.0.1\% when trained on GLIDE. Furthermore, when trained on VQDM and BigGAN, CAMME surpasses DCT by 10.23\% and 4.22\%, respectively. These gains are driven by CAMME's multimodal design, which goes beyond frequency patterns to integrate visual and text-based features, enabling a more nuanced understanding of the data. This results in greater cross-domain resilience compared to single-modal methods like UnivCLIP and DCT, enhancing CAMME's transferability and generalization performance on unseen target domains. Overall, when averaging across all transfer tasks (last column), CAMME achieves an F1 score of 77.71\%, outperforming the next best method, DCT, by 12.56\%.

\subsubsection{Face Dataset} \label{sec:face_results}
The baseline methods show poor performance on unseen domains (inter-domain), with DE-FAKE achieving an average F1 score of only 23.97\% on CelPG, 2.67\% on CelStar, and just 0.5\% on CelGlow. UnivCLIP performs well in source-only (intra-domain) settings (e.g., 98.06\% on CelGlow) but suffers a significant drop in average cross-domain (inter-domain) scores (e.g., an inter-domain F1 score of 36.77\% on CelGlow). DCT achieves competitive results on intra-domain settings, such as 99.95\% on CelStar, due to its frequency-based analysis; however, it underperforms in domains requiring visual-semantic understanding, trailing CAMME on complex datasets like YouFace. The YouFace dataset combines real images with varying lighting, poses, and expressions from YouTube-Frame, along with advanced reenactment manipulations from Face2Face, posing unique challenges for frequency-focused methods like DCT.

In contrast, CAMME consistently outperforms the baselines in inter-domain tasks, achieving top average F1 scores of 69.36\% on CelPG, 66.67\% on CelStar, and 66.62\% on CelGlow, representing an improvement of 27.21\% over the current state-of-the-art method. Single-modal methods such as UnivConv2B and DCT also struggle with the subtle generative artifacts introduced by StarGAN \cite{choiStarGANUnifiedGenerative2018} and Face2Face \cite{thiesFace2FaceRealtimeFace2020}. Specifically, UnivConv2B and DCT attain 0\% and 49.01\% on CelStar, and 40.35\% and 60.61\% on YouFace, respectively. The overall performance of the proposed CAMME, averaged across all tasks (last column), is 66.46\%, surpassing the next best method, UnivCLIP, by 13.25\%.

In the last row of Table \ref{table:face_results_simple}, we present the results of training on the YouFace dataset as the source domain. The findings indicate that the model trained on YouFace performs worse than the current baseline. Upon further analysis, we identified an issue originating from the prompt generation step of BLIP \cite{liBLIPBootstrappingLanguageImage2022}. Specifically, in the YouTube-Frame dataset, many frames are derived from the same individual, yet the BLIP model generates prompts describing entirely unrelated objects. This mismatch causes inaccurate semantic representations in the text embeddings, confusing the model and degrading performance on other datasets. Further analysis and examples of these failures are in Appendix \ref{app:fail_cases}.

\begin{table*}[!t]
    \centering{}\caption{Ablation study of different embedding aggregation methods (intra-domain performance shown in bold blue text).} \label{table:embedding_aggregation}\resizebox{1.0\textwidth}{!}{
    \begin{tabular}{|c|c|c|c|c|c|c|c|c|}
    \hline
    Methods &\diagbox{Source}{Target} & SD V1.5 & ADM & GLIDE & VQDM & BigGAN & \textbf{\makecell{Inter-domain\\Average (IA)}} & \textbf{Average of IA}\tabularnewline
    \hline
     & SD V1.5 & \textbf{\textcolor{blue}{88.76}} & 71.61 & 73.75 & 69.21 & 86.20 & 77.71 & \tabularnewline
    Element-wise Addition & ADM & 67.11 & \textbf{\textcolor{blue}{98.09}} & 65.98 & 94.83 & 65.46 & 58.68 & 56.53\tabularnewline
     & VQDM & 66.39 & 86.75 & 66.12 & \textbf{\textcolor{blue}{98.54}} & 65.53 & 33.20 & \tabularnewline
    \hline
     & SD V1.5 & \textbf{\textcolor{blue}{86.82}} & 72.61 & 62.75 & 69.74 & 80.33 & 71.36 & \tabularnewline
    Concatenation & ADM & 66.25 & \textbf{\textcolor{blue}{97.53}} & 65.42 & 94.02 & 65.18 & 72.72 & 71.57\tabularnewline
     & VQDM & 66.26 & 84.14 & 66.40 & \textbf{\textcolor{blue}{98.51}} & 65.72 & 70.63 & \tabularnewline
    \hline
     & SD V1.5 & \textbf{\textcolor{blue}{87.64}} & 74.13 & 63.21 & 71.65 & 81.78 & 72.69 & \tabularnewline
    Average Pooling & ADM & 64.88 & \textbf{\textcolor{blue}{94.07}} & 65.45 & 94.07 & 64.88 & 72.32 & 71.94\tabularnewline
     & VQDM & 65.81 & 86.23 & 65.64 & \textbf{\textcolor{blue}{98.08}} & 65.53 & 70.8 & \tabularnewline
    \hline
     & SD V1.5 & \textbf{\textcolor{blue}{92.87}} & 73.87 & 72.76 & 70.66 & 86.07 & 75.84 & \tabularnewline
    Learnable Weighted Average & ADM & 66.54 & \textbf{\textcolor{blue}{97.89}} & 66.37 & 96.58 & 65.43 & 73.73 & 73.58\tabularnewline
     & VQDM & 66.04 & 87.36 & 65.96 & \textbf{\textcolor{blue}{97.90}} & 65.27 & 71.16 & \tabularnewline
    \hline
     & SD V1.5 & \textbf{\textcolor{blue}{99.49}} & 66.47 & 74.76 & 68.35 & 72.03 & 70.40 & \tabularnewline
    Multi-Head Cross-Attention & ADM & 66.19 & \textbf{\textcolor{blue}{99.47}} & 95.65 & 98.97 & 99.26 & 90.02 & \textbf{81.63}\tabularnewline
     & VQDM & 66.73 & 94.12 & 84.60 & \textbf{\textcolor{blue}{99.79}} & 92.39 & 84.46 & \tabularnewline
    \hline
    \end{tabular}}
    \end{table*}

  \begin{table*}[t]
        \centering{}\caption{Ablation study of different embedding combinations (intra-domain performance shown in bold blue text).}\label{table:embedding_combination}\resizebox{1.0\textwidth}{!}{
        \begin{tabular}{|c|c|c|c|c|c|c|c|c|}
        \hline
        Methods & \backslashbox[20mm]{Source}{Target} & SD V1.5 & ADM & GLIDE & VQDM & BigGAN & \textbf{\makecell{Inter-domain\\Average (IA)}} & \textbf{Average of IA}\tabularnewline


        \hline
        \multirow{2}{*}{Visual and Text} & BigGAN & 66.62 & 68.02 & 68.68 & 71.65 & \textbf{\textcolor{blue}{99.94}} & 68.74 & \multirow{2}{*}{79.70}\tabularnewline
         & ADM & 66.35 & \textbf{\textcolor{blue}{99.44}} & 97.97 & 98.85 & 99.41 & 90.65 & \tabularnewline
        \hline
        \multirow{2}{*}{Visual and Frequency} & BigGAN & 67.58 & 69.10 & 70.29 & 71.07 & \textbf{\textcolor{blue}{92.32}} & 69.51 & \multirow{2}{*}{79.42}\tabularnewline
         & ADM & 66.44 & \textbf{\textcolor{blue}{99.35}} & 93.00 & 98.82 & 99.03 & 89.32 & \tabularnewline
        \hline
        \multirow{2}{*}{Text and Frequency} & BigGAN & 65.13 & 66.22 & 65.35 & 65.88 & \textbf{\textcolor{blue}{97.48}} & 65.65 & \multirow{2}{*}{68.78}\tabularnewline
         & ADM & 65.40 & \textbf{\textcolor{blue}{96.55}} & 65.55 & 92.13 & 64.55 & 71.91 & \tabularnewline
        \hline
        \multirow{2}{*}{Visual, Text and Frequency} & BigGAN & 67.08 & 70.48 & 74.94 & 78.40 & \textbf{\textcolor{blue}{99.76}} & 72.73 & \multirow{2}{*}{\textbf{81.38}}\tabularnewline
         & ADM & 66.19 & \textbf{\textcolor{blue}{99.47}} & 95.65 & 98.97 & 99.26 & 90.02 & \tabularnewline
        \hline
        \end{tabular}}
        \end{table*}

\subsection{Analysis}

\subsubsection{Embedding Combination Methods} \label{sec:embedding_combination}

We evaluate different strategies for combining visual, textual, and frequency embeddings in our proposed method. Specifically, we compare the performance of the following five approaches:

\begin{enumerate}[label=(\alph*)]
    \item \textit{Element-wise Addition}: The embeddings are combined by summing them element-wise: $\bar{\textbf{e}} = \textbf{e}_I + \textbf{e}_T + \textbf{e}_F$.
    \item \textit{Concatenation}: The embeddings are concatenated along the channel dimension: $\bar{\textbf{e}} = [\textbf{e}_I, \textbf{e}_T, \textbf{e}_F]$.
    \item \textit{Average Pooling}: The embeddings are combined by averaging their values across the channel dimension: $\bar{\textbf{e}} = \frac{1}{3}(\textbf{e}_I + \textbf{e}_T + \textbf{e}_F)$.
    \item \textit{Learnable Weighted Average}: Trainable weights $\textbf{w} \in \mathbb{R}^{1\times3}$ are initialized, scaled via the softmax function, and learned during training: $\bar{\textbf{e}} = \text{softmax}(\textbf{w}) \cdot [\textbf{e}_I, \textbf{e}_T, \textbf{e}_F]$.
    \item \textit{Multi-Head Cross-Attention}: Our proposed method, which uses a cross-attention mechanism to dynamically aggregate information from the embeddings.
\end{enumerate}

For methods (a), (b), (c), and (d), we then apply a linear layer to the combined embedding $\bar{\textbf{e}}$ for classification. The experimental results are presented in Table \ref{table:embedding_aggregation}.

Our proposed model (CAMME) with multi-head cross-attention achieves the highest average F1 score of 81.63\%, outperforming the second-best method, \textit{Learnable Weighted Average}, by 8.05\%. This result demonstrates the effectiveness of our method, showing that even a single-head configuration can capture interactions between different embeddings and leverage their complementary information to enhance deepfake detection performance.

\begin{table*}
    \centering
    \caption{Average F1 scores for varying intensities of natural perturbations on manipulated images from GLIDE dataset.}
    \label{tab:abla_natural_pertub}
    \begin{tabular}{|c|c|c|c|c|c|c|}
    \hline
    Methods & Unperturbed & Gaus. Noise & Gaus. Blur & JPEG Comp. & Sharpening & Color Jittering\tabularnewline
    (FID values)  & - & 2.73 & 4.40 & 3.23 & 3.17 & 6.46\tabularnewline
    \hline
    DE-FAKE & 98.86 & 82.99 & 80.84 & 56.04 & 98.18 & 97.87\tabularnewline
    DCT & 99.17 & 23.86 & 75.75 & 32.64 & 88.69 & 61.13\tabularnewline
    UnivCLIP & 97.63 & 89.39 & 93.12 & 93.45 & 95.35 & 89.69\tabularnewline
    UnivConv2B & 99.97 & 65.44 & 82.17 & 41.85 & 97.00 & 97.54\tabularnewline
    \hline
    \textbf{CAMME} & \textbf{99.65} & \textbf{91.36} & \textbf{96.05} & \textbf{99.12} & \textbf{98.70} & \textbf{97.91}\tabularnewline
    \hline
\end{tabular}
\end{table*}

\begin{figure*}
\begin{centering}
\resizebox{0.8\textwidth}{!}{%
\includegraphics[width=\textwidth]{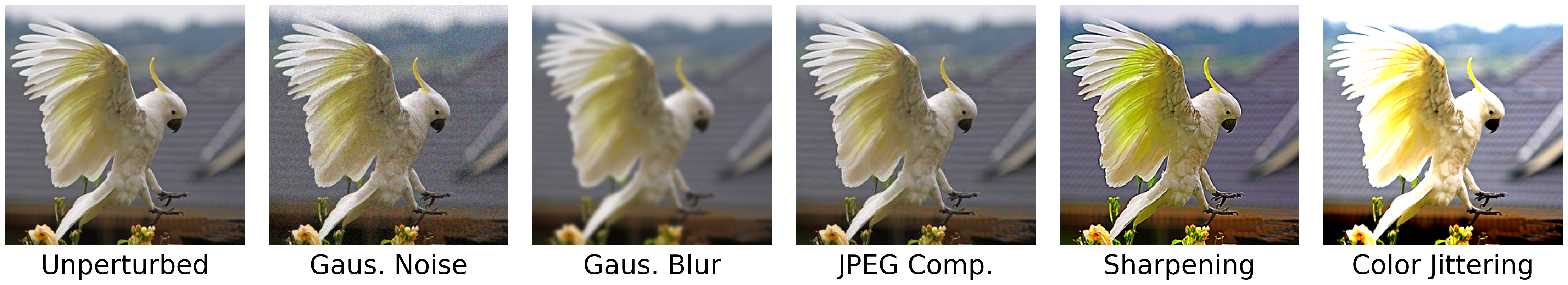}
}
\caption{Examples of various natural perturbations analyzed in our study.}
\label{fig:natural_perturbation}
\par\end{centering}
\end{figure*}

\subsubsection{Input Embedding Scenario Analysis}
In this section, we investigate the impact of different input embeddings on the performance of the cross-attention mechanism. Specifically, we omit one token embedding from $\textbf{e}$ in Equation (\ref{eq:token_embedding}) and conduct experiments on various scenarios: (i) \textit{Visual and Text}, (ii) \textit{Visual and Frequency}, (iii) \textit{Text and Frequency}, and (iv) \textit{Visual, Text, and Frequency} (proposed method). The results are presented in Table \ref{table:embedding_combination}.

From the table, it is evident that excluding visual embeddings leads to significantly lower performance, as seen in the \textit{Text and Frequency} combination, which achieves an average F1 score of 68.78\%. The inclusion of frequency or text embeddings proves beneficial, particularly when combined with visual data, as demonstrated by the high scores in the \textit{Visual and Text} (79.70\%) and \textit{Visual and Frequency} (79.42\%) combinations, respectively. Our proposed method, which integrates all three embeddings, achieves the highest average F1 score of 81.38\%, surpassing the second-best combination, \textit{Visual and Text}, by 1.68\%. This combination provides the model with the most comprehensive representation, capturing visual, semantic, and spectral information, which improves overall detection accuracy.

\subsubsection{Natural Perturbations}
This section evaluates the proposed model's performance under traditional image transformations, including Gaussian noise, Gaussian blur, JPEG compression, sharpening, and color jittering. The GLIDE \cite{nicholGLIDEPhotorealisticImage2022} dataset was chosen due to its high detection accuracy across all baselines, enabling an analysis of performance drops after applying these transformations. All test images from the test set of GLIDE were used, with hyperparameters adjusted for different intensity levels. The average F1 score was reported, as it provides a balanced measure of precision and recall. In this scenario, minimizing false negatives (fake images classified as real) and false positives (real images classified as fake) is crucial. Natural perturbations introduced by image editing tools or apps may cause misclassifications, especially if real images are mistakenly flagged as fake. Note that these transformations are not considered fake manipulations as they do not alter the intrinsic content of the images. Figure~\ref{fig:natural_perturbation} illustrates an example of these perturbations before and after application. In each figure, we compute the FID score \cite{heuselGANs2017} to assess the quality of natural perturbations by comparing their feature distributions to real images. Higher FID scores indicate more significant perturbations. Results are summarized in Table \ref{tab:abla_natural_pertub}.
\paragraph{Gaussian Noise}
Gaussian noise involves adding random pixel variations based on a Gaussian distribution. In this experiment, the mean was set to 0, and the standard deviation (\(\sigma\)) was varied as \(\{0.01, 0.015, 0.02\}\).

\paragraph{Gaussian Blur}
Gaussian blur reduces noise or smooths edges by applying a Gaussian function:
\begin{equation}
G(x, y) = \frac{1}{2\pi\sigma^2}\exp\left(-\frac{x^2 + y^2}{2\sigma^2}\right),
\end{equation}
where \(x\) and \(y\) are pixel coordinates, and \(\sigma\) determines blur intensity. Values of \(\sigma\) were tested from \(\{0.5, 1.0, 1.5\}\).

\paragraph{JPEG Compression}
JPEG compression's impact on deepfake detection was evaluated by varying the quality factor, representing compression intensity, across \(\{30, 60, 90\}\). Higher quality factors result in less compression and better image fidelity \cite{Lau2003,s23073400}.

\paragraph{Sharpening}
Sharpening enhances image details by increasing the contrast between adjacent pixels. Experiments were conducted with sharpening factors of \(\{1.0, 1.5, 2.0\}\).

\paragraph{Color Jittering}
Color jittering, a common augmentation technique, alters image properties like brightness, contrast, saturation, and hue. This experiment focused on brightness and contrast. We used tune the same values for both from \(\{1.0, 1.5, 2.0\}\).

Table \ref{tab:abla_natural_pertub} highlights the robustness of different methods under various natural perturbations. CAMME consistently outperforms other methods, maintaining high F1 scores across all perturbations, including Gaussian noise (91.36\%), Gaussian blur (96.05\%), and JPEG compression (99.12\%). This demonstrates CAMME's ability to effectively handle transformations that degrade image quality, as reflected in the FID values. In contrast, methods like DCT showed significant vulnerabilities (23.86\% on Gaussian noise), while UnivCLIP and UnivConv2B performed competitively in certain scenarios, such as Gaussian blur (93.12\%) and sharpening (97.00\%), but fell short under more challenging transformations. The FID values indicate that transformations like color jittering (6.46) or Gaussian blur (4.40) introduce moderate perceptual degradation. However, the high F1 scores achieved by CAMME under these perturbations demonstrate its robustness to natural image transformations, highlighting its potential for detecting deepfakes in real-world scenarios.

\subsubsection{Resilience to Adversarial Attacks}
In this section, we evaluate the robustness of CAMME and other deepfake detectors against adversarial perturbations. Two widely used adversarial attacks, Fast Gradient Sign Method (FGSM) \cite{goodfellowExplainingHarnessingAdversarial2015} and Projected Gradient Descent (PGD) \cite{madryDeepLearningModels2019}, are applied to generate adversarial examples from clean fake images. The GLIDE test dataset, consisting of 1,700 fake images, is used for this evaluation. Figure \ref{fig:adversarial_examples} in Appendix \ref{sec:adversarial_examples} illustrates examples of fake images before and after the adversarial attacks.

\begin{table}[t]
    \caption{Robustness performance (F1 score) against adversarial attacks.} \label{table:adversarial_results}
    \resizebox{1.0\columnwidth}{!}{%
    \begin{tabular}{|c|c|c|c|}
    \hline
    \multirow{2}{*}{Methods} & Natural & FGSM Attack & PGD Attack\\
    & Accuracy & Accuracy & Accuracy \\
    \hline
    DE-FAKE & 98.86 & 0.00 & 0.00\tabularnewline
    UnivCLIP & 97.63 & 43.04 & 21.43\tabularnewline
    UnivConv2B & \textbf{99.97} & 66.56 & 66.68\tabularnewline
    DCT & 99.20 & 2.31 & 0.00 \tabularnewline
    \textbf{CAMME} & 99.65 & \textbf{96.14} & \textbf{89.01}\tabularnewline
    \hline
    \end{tabular}}
\end{table}

\paragraph{Attack Selection.} FGSM is a one-step attack that perturbs the input image in the direction of the gradient of the loss function with respect to the input. In contrast, PGD generates adaptive adversaries by iteratively refining perturbations using the model's gradients, resulting in a more tailored and robust attack against the target model. Both are widely studied white-box attacks, assuming the attacker has full access to the detector model's architecture and parameters. The formula for FGSM is given by:
\begin{equation}
    x_{\text{adv}} = x + \epsilon \cdot \text{sign}(\nabla_x \mathcal{L}(f(x;\phi), y))
\end{equation}
where $x$ is the input image, $x_{\text{adv}}$ is the adversarial image, $\epsilon$ is the perturbation strength, $\mathcal{L}$ is the loss function, $f$ is the detector model parametrized by $\phi$, and $y$ is the target label. On the other hand, PGD is a multi-step iterative attack that optimizes the objective function over multiple steps and projects each step ($t$) back within the $\epsilon$-ball $B_{\epsilon}(x)=\{x': \lVert x' - x \rVert_{\infty } \leq \epsilon\}$.
\begin{equation}
    x^{t+1}_{\text{adv}} = \text{Proj}_{B_{\epsilon}(x)} \left( x^t_{\text{adv}} + \alpha \cdot \text{sign}(\nabla_x \mathcal{L}(f(x^t_{\text{adv}};\phi), y)) \right),
\end{equation}
where $x^{0}_{\text{adv}} = x$. To run these attacks, for FGSM, we set $\epsilon = 0.1$. For PGD, we set $\epsilon = 0.1$, $\alpha = 0.01$ and $t = 20$.

\begin{table*}[!t]
    \centering{}\caption{Ablation study on varying the number of attention heads (intra-domain
    performance shown in bold blue text).}\label{table:num_heads}
    \begin{tabular}{|c|c|c|c|c|c|c|c|}
    \hline
    Number of Heads & \diagbox{Source}{Target} & SD V1.5 & ADM & GLIDE & VQDM & BigGAN & \textbf{\makecell{Inter-domain\\Average (IA)}}\tabularnewline
    \hline
    \multirow{1}{*}{$1$} & BigGAN & 66.83 & 68.97 & 69.93 & 70.48 & \textbf{\textcolor{blue}{99.82}} & 69.05\tabularnewline
    \hline
    \multirow{1}{*}{$2$} & BigGAN & 66.43 & 69.39 & 69.20 & 71.32 & \textbf{\textcolor{blue}{99.65}} & 69.09\tabularnewline
    \hline
    \multirow{1}{*}{$4$} & BigGAN & 66.47 & 69.23 & 67.95 & 70.13 & \textbf{\textcolor{blue}{99.82}} & 68.45\tabularnewline
    \hline
    \multirow{1}{*}{$8$} & BigGAN & 67.08 & 70.48 & 74.94 & 78.40 & \textbf{\textcolor{blue}{99.76}} & \textbf{72.73}\tabularnewline
    \hline
    \multirow{1}{*}{$16$} & BigGAN & 67.68 & 71.32 & 73.77 & 73.59 & \textbf{\textcolor{blue}{99.35}} & 71.59\tabularnewline
    \hline
    \end{tabular}
    \end{table*}
\begin{table*}[!t]
\centering
\caption{Detection transferability performance across different input image sizes (intra-domain performance highlighted in bold blue text).}
\label{table:image_size}
\renewcommand{\arraystretch}{1.1}
\setlength{\tabcolsep}{4pt} 
\begin{tabularx}{\textwidth}{|c|X|c|c|c|c|c|c|c|}
\hline
\textbf{Methods} & \diagbox{Source}{Target} & \textbf{SD V1.5} & \textbf{ADM} & \textbf{GLIDE} & \textbf{VQDM} & \textbf{BigGAN} & \textbf{\makecell{Inter-domain \\Avg (IA)}} & \textbf{\makecell{Avg IA}} \\
\hline
\multirow{2}{*}{$32\times32$} & SD V1.5 & \textbf{\textcolor{blue}{91.37}} & 66.60 & 71.08 & 68.98 & 67.46 & 68.53 & \multirow{2}{*}{68.12} \\
 & BigGAN & 67.69 & 67.20 & 67.62 & 68.30 & \textbf{\textcolor{blue}{80.66}} & 67.70 & \\
\hline
\multirow{2}{*}{$64\times64$} & SD V1.5 & \textbf{\textcolor{blue}{94.32}} & 66.51 & 71.60 & 68.27 & 66.61 & 68.25 & \multirow{2}{*}{69.51} \\
 & BigGAN & 67.45 & 70.87 & 77.78 & 66.94 & \textbf{\textcolor{blue}{94.48}} & 70.76 & \\
\hline
\multirow{2}{*}{$128\times128$} & SD V1.5 & \textbf{\textcolor{blue}{95.28}} & 65.70 & 78.79 & 67.95 & 66.68 & 69.78 & \multirow{2}{*}{68.26} \\
 & BigGAN & 66.67 & 66.67 & 66.83 & 66.77 & \textbf{\textcolor{blue}{99.73}} & 66.74 & \\
\hline
\multirow{2}{*}{$224\times224$} & SD V1.5 & \textbf{\textcolor{blue}{97.80}} & 66.57 & 81.01 & 66.12 & 67.30 & 70.25 & \multirow{2}{*}{70.98} \\
 & BigGAN & 68.20 & 67.44 & 84.14 & 67.05 & \textbf{\textcolor{blue}{98.27}} & 71.71 & \\
\hline
\multirow{2}{*}{$512\times512$} & SD V1.5 & \textbf{\textcolor{blue}{98.33}} & 73.82 & 84.97 & 80.13 & 72.13 & 77.76 & \multirow{2}{*}{\textbf{74.08}} \\
 & BigGAN & 66.29 & 66.59 & 74.33 & 74.33 & \textbf{\textcolor{blue}{99.17}} & 70.39 & \\
\hline
\end{tabularx}
\end{table*}

\paragraph{Results.} Table \ref{table:adversarial_results} compares the robustness of various methods against selected attacks. The results demonstrate the strong performance of our proposed model, CAMME, in both generalization and resilience to adversarial perturbations. CAMME achieves a natural accuracy of 99.65\%, showcasing its robust generalization on clean images, while UnivConv2B slightly outperforms it with 99.97\%. However, CAMME excels under adversarial settings. Against FGSM attacks, CAMME achieves the highest F1 score of 96.14\%, significantly surpassing UnivConv2B at 66.56\%. Other methods, such as DE-FAKE and DCT, fail completely under FGSM attacks, with F1 scores of 0\%.

For the more challenging PGD attacks, CAMME achieves an F1 score of 89.01\%, representing a substantial improvement over UnivConv2B, the second-best method, which scores 66.68\%. Other methods, including UnivCLIP and DCT, show limited or no robustness, achieving 21.43\% and 0.00\%, respectively. These results highlight the limitations of baseline methods that rely on limited modalities or simple embeddings, making them vulnerable to adversarial perturbations. In contrast, CAMME's integration of visual, textual, and frequency-domain embeddings enables it to capture richer input representations, significantly enhancing its robustness.

Overall, CAMME demonstrates strong generalization on clean images and exceptional resilience to adversarial attacks, outperforming existing methods across all scenarios. Its ability to leverage multi-modal embeddings and cross-attention mechanisms plays a crucial role in achieving these results, highlighting its effectiveness as a robust deepfake detection model.

\subsubsection{Single-Head vs. Multi-Head Cross-Attention}
The number of attention heads is a key hyperparameter in transformer-based models, as it governs the model's ability to capture complex interactions between embeddings. Table \ref{table:num_heads} shows the performance of our model across inter-domain tasks as the number of attention heads $H$ is varied in $\{1, 2, 4, 8, 16\}$. The results (in F1 score) indicate a consistent improvement in inter-domain performance as the number of heads increases, reaching the highest average performance of 72.73\% when $H = 8$. For instance, at $H = 1$, the inter-domain average (IA) is 69.05\%, which steadily increases to 69.09\% at $H = 2$, 68.45\% at $H = 4$, and peaks at $H = 8$ with 72.73\%. However, a slight decline to 71.59\% is observed when the number of heads is increased to 16.
The optimal performance at $H = 8$ can be attributed to the model's ability to strike a balance between adequately capturing the relationships across embeddings and maintaining computational efficiency. With fewer heads, such as $H = 1$, the model struggles to capture enough diverse interactions, leading to lower detection accuracy. On the other hand, while increasing the number of heads to $H = 16$ adds more capacity to model interactions, it introduces redundancy and potential overfitting, slightly degrading performance. Based on these results, we recommend setting the number of attention heads to $H = 8$, as it provides the best balance between complexity and performance, effectively capturing multi-modal interactions and enhancing detection transferability.


\subsubsection{Impact of Input Image Size}

This section analyzes the effect of input image size on the detection performance of our proposed model. Table \ref{table:image_size} presents results across resolutions: $32 \times 32$, $64 \times 64$, $128 \times 128$, $224 \times 224$, and $512 \times 512$, using images from BigGAN and Stable Diffusion (SD V1.5). Detection performance improves with larger resolutions, as higher spatial details enable better identification of subtle artifacts. For BigGAN, the best inter-domain performance is achieved at $224 \times 224$ with an F1 score of 71.71\% and an intra-domain F1 score of 98.27\%. However, at $512 \times 512$, inter-domain performance slightly decreases to 70.39\%, potentially due to redundancy in spatial details. In contrast, SD V1.5 consistently benefits from increased resolution, peaking at $512 \times 512$ with an inter-domain F1 score of 77.76\% and an intra-domain F1 score of 98.33\%. The lowest performance for both models is observed at $32 \times 32$, where critical spatial details are lost, resulting in F1 scores of 67.70\% for BigGAN and 68.53\% for SD V1.5. In summary, the highest inter-domain performance of 74.08\% is achieved at $512 \times 512$, demonstrating the importance of higher resolutions. Larger inputs enhance the model's ability to capture spatial and spectral patterns, improving the effectiveness of frequency transformations like the Discrete Cosine Transform (DCT) and enabling more robust, transferable embeddings that enhance detection performance across diverse domains.


\section{Conclusion}
This paper introduces CAMME (Cross-Attention Multi-Modal Embeddings), a framework addressing the critical challenge of cross-domain generalization in deepfake detection. By dynamically integrating visual, textual, and frequency-domain features through multi-head cross-attention, CAMME establishes robust decision boundaries that effectively generalize to unseen generative architectures. Our extensive experiments across 41 transfer tasks demonstrate significant performance improvements over state-of-the-art methods: 12.56\% on natural scenes and 13.25\% on facial deepfakes.

CAMME's multi-modal approach provides inherent resilience against both natural and adversarial perturbations, maintaining over 91\% accuracy under image transformations and achieving 96.14\% and 89.01\% F1 scores against FGSM and PGD attacks, respectively. These results confirm that cross-modal integration through attention mechanisms effectively mitigates the domain-specificity limitations of previous approaches.

The demonstrated generalization capabilities and robustness position CAMME as an effective countermeasure against the rapidly evolving landscape of generative AI technologies. Future work could explore extending this approach to video deepfakes and investigating self-supervised adaptation techniques for emerging generative architectures.

\bibliographystyle{unsrt}
\bibliography{references.bib}

\appendix
\begin{figure*}[t!]
  \centering
  \begin{tabular}{*{9}{c@{\hspace{2pt}}}} 
    \multicolumn{5}{c}{\fontsize{10}{10}\selectfont Natural Scene Images} & 
    \multicolumn{1}{c}{} & 
    \multicolumn{3}{c}{\fontsize{10}{10}\selectfont Face Images} \\

    \includegraphics[width=1.6cm, height=1.6cm]{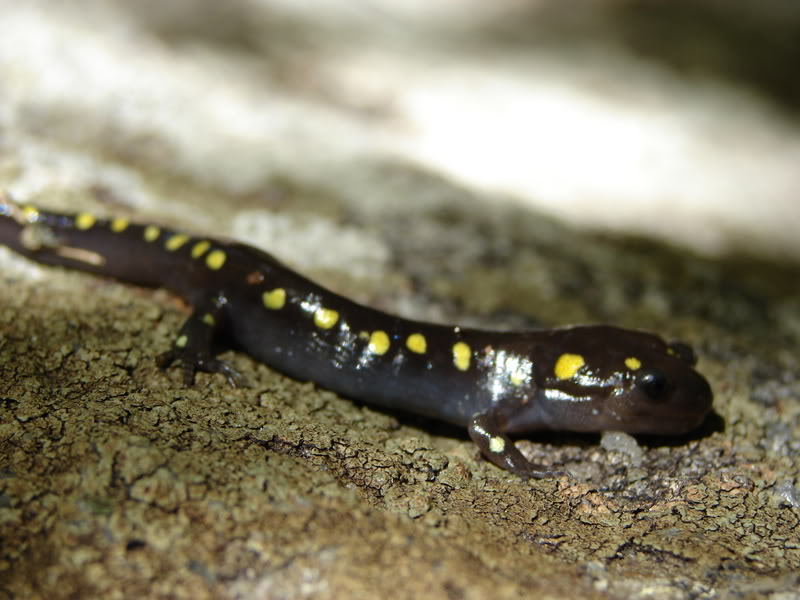} &
    \includegraphics[width=1.6cm, height=1.6cm]{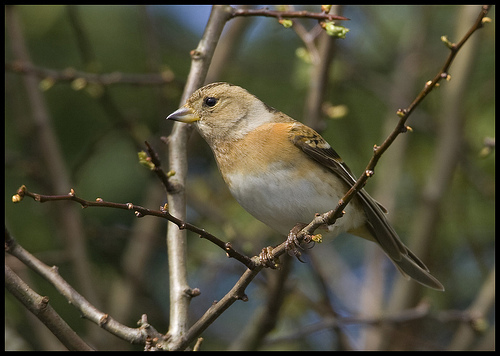} &
    \includegraphics[width=1.6cm, height=1.6cm]{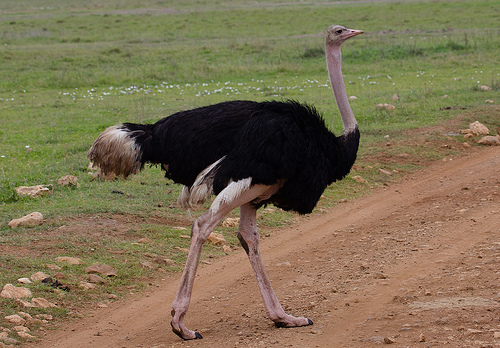} &
    \includegraphics[width=1.6cm, height=1.6cm]{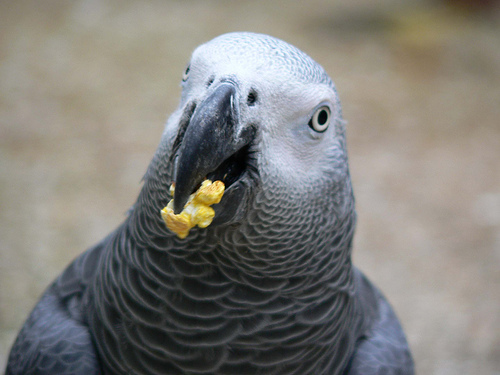} &
    \includegraphics[width=1.6cm, height=1.6cm]{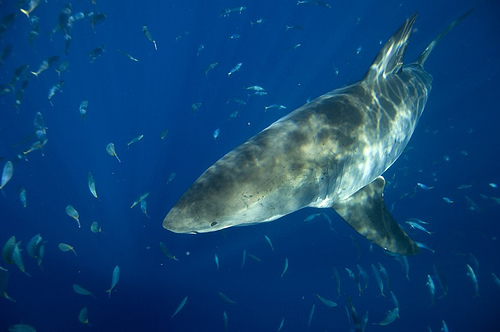} &
    \includegraphics[width=1.6cm, height=1.6cm]{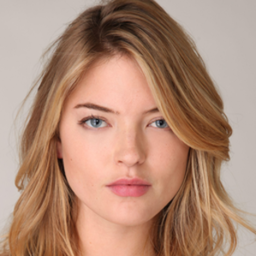} &
    \includegraphics[width=1.6cm, height=1.6cm]{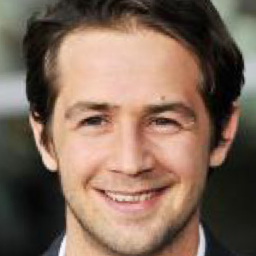} &
    \includegraphics[width=1.6cm, height=1.6cm]{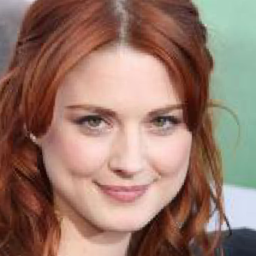} &
    \includegraphics[width=1.6cm, height=1.6cm]{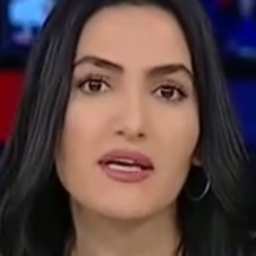} \\

    \includegraphics[width=1.6cm, height=1.6cm]{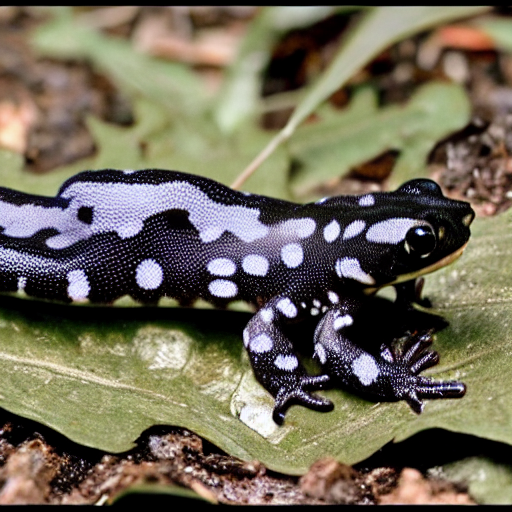} &
    \includegraphics[width=1.6cm, height=1.6cm]{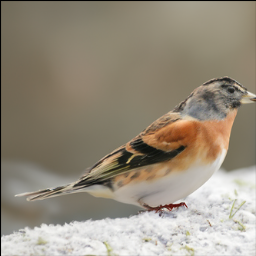} &
    \includegraphics[width=1.6cm, height=1.6cm]{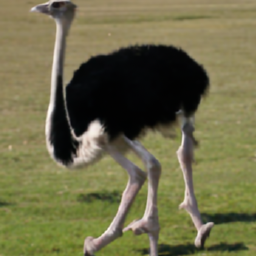} &
    \includegraphics[width=1.6cm, height=1.6cm]{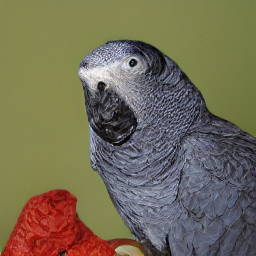} &
    \includegraphics[width=1.6cm, height=1.6cm]{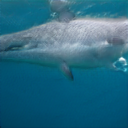} &
    \includegraphics[width=1.6cm, height=1.6cm]{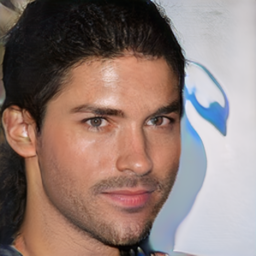} &
    \includegraphics[width=1.6cm, height=1.6cm]{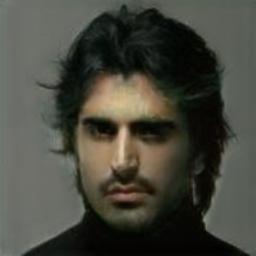} &
    \includegraphics[width=1.6cm, height=1.6cm]{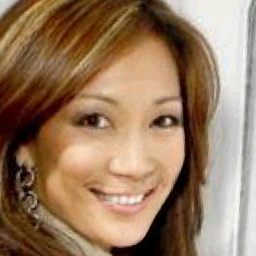} &
    \includegraphics[width=1.6cm, height=1.6cm]{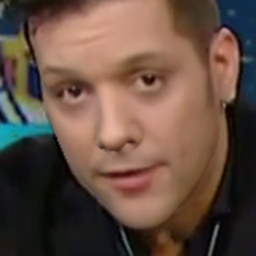} \\

    {\fontsize{8}{8}\selectfont (a) SD V1.5} & 
    {\fontsize{8}{8}\selectfont (b) ADM} & 
    {\fontsize{8}{8}\selectfont (c) GLIDE} &
    {\fontsize{8}{8}\selectfont (d) VQDM} & 
    {\fontsize{8}{8}\selectfont (e) BigGAN} &
    {\fontsize{8}{8}\selectfont (f) CelPG} & 
    {\fontsize{8}{8}\selectfont (g) CelStar} & 
    {\fontsize{8}{8}\selectfont (h) CelGlow} &
    {\fontsize{8}{8}\selectfont (i) YouFace} \\
  \end{tabular}
  \caption{Sample image pairs (real/fake) from different generative models. Top: real images; bottom: synthetic. Left: natural scenes; Right: face generators.}
  \label{fig:combined_real_and_fake}
\end{figure*}

\begin{figure*}
    \resizebox{\textwidth}{!}{%
    \begin{tabular}[t]{cccccc}
    \includegraphics{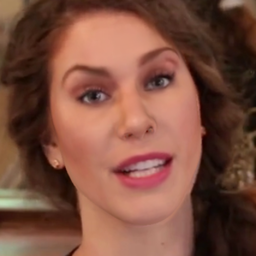} &
    \includegraphics{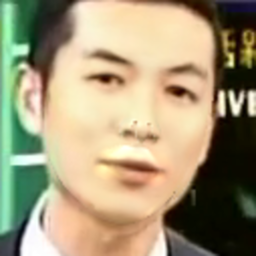} &
    \includegraphics{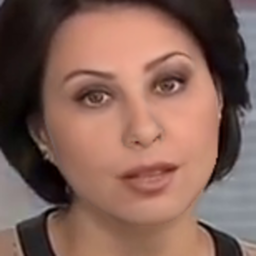} &
    \includegraphics{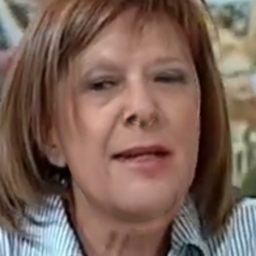} &
    \includegraphics{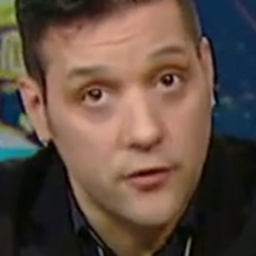} &
    \includegraphics{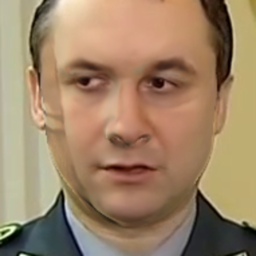} \\
    
    \begin{varwidth}{0.65\linewidth}
    \centering
    \huge a close up of a woman \\ with a necklace on a table
    \end{varwidth} &
    \begin{varwidth}{0.65\linewidth}
    \centering
    \huge arafed asian man in a suit \\ and tie standing in front of a tv
    \end{varwidth} &
    \begin{varwidth}{0.65\linewidth}
    \centering
    \huge a close up of a woman \\ with a necklace on a tv screen
    \end{varwidth} &
    \begin{varwidth}{0.65\linewidth}
    \centering
    \huge a close up of a woman \\ with a tie on a couch
    \end{varwidth} &
    \begin{varwidth}{0.65\linewidth}
    \centering
    \huge arafed man in a black jacket \\ and tie sitting in front of a tv
    \end{varwidth} &
    \begin{varwidth}{0.65\linewidth}
    \centering
    \huge arafed man in uniform \\ with a tie and a tie around his neck
    \end{varwidth}
    \end{tabular}}
    \caption{Investigation of misalignment between visual and text content in the YouFace dataset.}
    \label{fig:fail_cases}
\end{figure*}

\section{Dataset Details} \label{app:dataset_details}

In this section, we describe the dataset split used in our study. The Natural Scene dataset comprises images generated by various models, including SD V1.5, ADM, GLIDE, VQDM, and BigGAN, while the Face dataset consists of different real-fake combinations: CelPG, CelStar, CelGlow, and YouFace. Table \ref{table:dataset_details} presents the dataset composition for both the Natural Scene and Face datasets.

Images in the Natural Scene dataset are grouped into 100 categories, specifically chosen from the first 100 categories in ImageNet \cite{dengImageNetLargeScaleHierarchical2009}. In contrast, the Face dataset features a diverse set of images that are not grouped into specific categories. For the Face dataset, we define the real and fake image combinations as follows:
\begin{itemize}
    \item \textit{CelPG}: CelebA-HQ (real) and PGGAN (fake)
    \item \textit{CelStar}: CelebA (real) and StarGAN (fake)
    \item \textit{CelGlow}: CelebA (real) and Glow (fake)
    \item \textit{YouFace}: Youtube-Frame (real) and Face2Face (fake)
\end{itemize}
These combinations ensure a balanced dataset with consistent characteristics across realistic and synthetic face pairs, enhancing robustness in deepfake detection. Figure \ref{fig:combined_real_and_fake} provide examples of real and fake images from the Natural Scene and Face datasets, respectively.

\section{Detailed Results on Natural Scene and Face Datasets} \label{app:full_results}
In this section, we present the full results for deepfake detection performance on the Natural Scene and Face datasets. Table \ref{table:natural_scene_full} provides the precision, recall, F1 score, and accuracy for each model on the Natural Scene dataset. Similarly, Table \ref{table:face_full} presents the performance metrics for the Face dataset. The inter-domain average (IA) and the average of IA are also included for each model. The results demonstrate the effectiveness of our proposed model in enhancing cross-domain transferability and robustness in deepfake detection, particularly in the challenging Face dataset.

\section{Investigation of Fail Cases} \label{app:fail_cases}
As discussed in Section \ref{sec:face_results}, we investigate the misalignment between visual and textual content in the YouFace dataset. The results are illustrated in Figure \ref{fig:fail_cases}. YouFace is a particularly challenging dataset, combining real images with varying lighting conditions, poses, and expressions from YouTube-Frame \cite{rosslerFaceForensicsLearningDetect2019}, alongside advanced reenactment manipulations from Face2Face \cite{thiesFace2FaceRealtimeFace2020}. Additionally, the YouTube-Frame dataset contains repetitive content, such as a person speaking in front of a camera across multiple frames of the same video. This repetition confuses BLIP \cite{liBLIPBootstrappingLanguageImage2022}, making it difficult to generate precise captions that accurately capture the semantic meaning of the images. For instance, the caption generated for the first image in Figure \ref{fig:fail_cases} is ``\textit{a woman with a necklace on a table}''. However, neither a necklace nor a table is present in the image.

Such inconsistencies lead to ineffective alignment and weakened interactions between textual, visual, and frequency embeddings in our proposed model. This misalignment—whether between visual and textual embeddings or between textual and frequency embeddings—causes a drop in detection performance on the YouFace dataset. A potential solution to this issue is to improve the quality of text embeddings by training a more robust BLIP model on a larger and more diverse dataset. Alternatively, exploring specialized image-to-text generation models tailored for facial content may offer a more effective alignment mechanism in the future.

\begin{table}
    \centering{}%
    \caption{Dataset composition for Natural Scene and Face Datasets. Each dataset
    lists the number of fake and real images used in the training, validation,
    and test sets.} \label{table:dataset_details}
    \resizebox{\columnwidth}{!}{
    \begin{tabular}{|c|c|c|c|c|}
    \hline 
    \multirow{2}{*}{Dataset Type} & \multirow{2}{*}{Datasets} & Train & Validation  & Test\tabularnewline
     &  & (Fake/Real) & (Fake/Real) & (Fake/Real)\tabularnewline
    \hline 
    \multirow{5}{*}{
        Natural Scene
    } & SD V1.5 & 12,898/12,900 & 1,600/1,600 & 1,699/1,700\tabularnewline
    \cline{2-5}
     & ADM & 12,837/12,837 & 1,591/1,591 & 1,692/1,692\tabularnewline
    \cline{2-5}
     & GLIDE & 12,900/12,900 & 1,600/1,600 & 1,700/1,700\tabularnewline
    \cline{2-5}
     & VQDM & 12,900/12,900 & 1,600/1,600 & 1,700/1,700\tabularnewline
    \cline{2-5}
     & BigGAN & 12,900/12,900 & 1,600/1,600 & 1,700/1,700\tabularnewline
    \hline 
    \multirow{4}{*}{Face} & CelPG & 8,000/8,000 & 1,000/1,000 & 1,000/1,000\tabularnewline
    \cline{2-5}
     & CelStar & 8,000/8,000 & 1,000/1,000 & 1,000/1,000\tabularnewline
    \cline{2-5}
     & CelGlow & 8,000/8,000 & 1,000/1,000 & 1,000/1,000\tabularnewline
    \cline{2-5}
     & YouFace & 8,000/8,000 & 1,000/1,000 & 1,000/1,000\tabularnewline
    \hline 
    \end{tabular}}
    \end{table}
    
\begin{figure}[t]
    \centering{}\resizebox{\columnwidth}{!}{%
    \fontsize{30}{30}\selectfont
    \begin{tabular}{c}
    \includegraphics[scale=1.0]{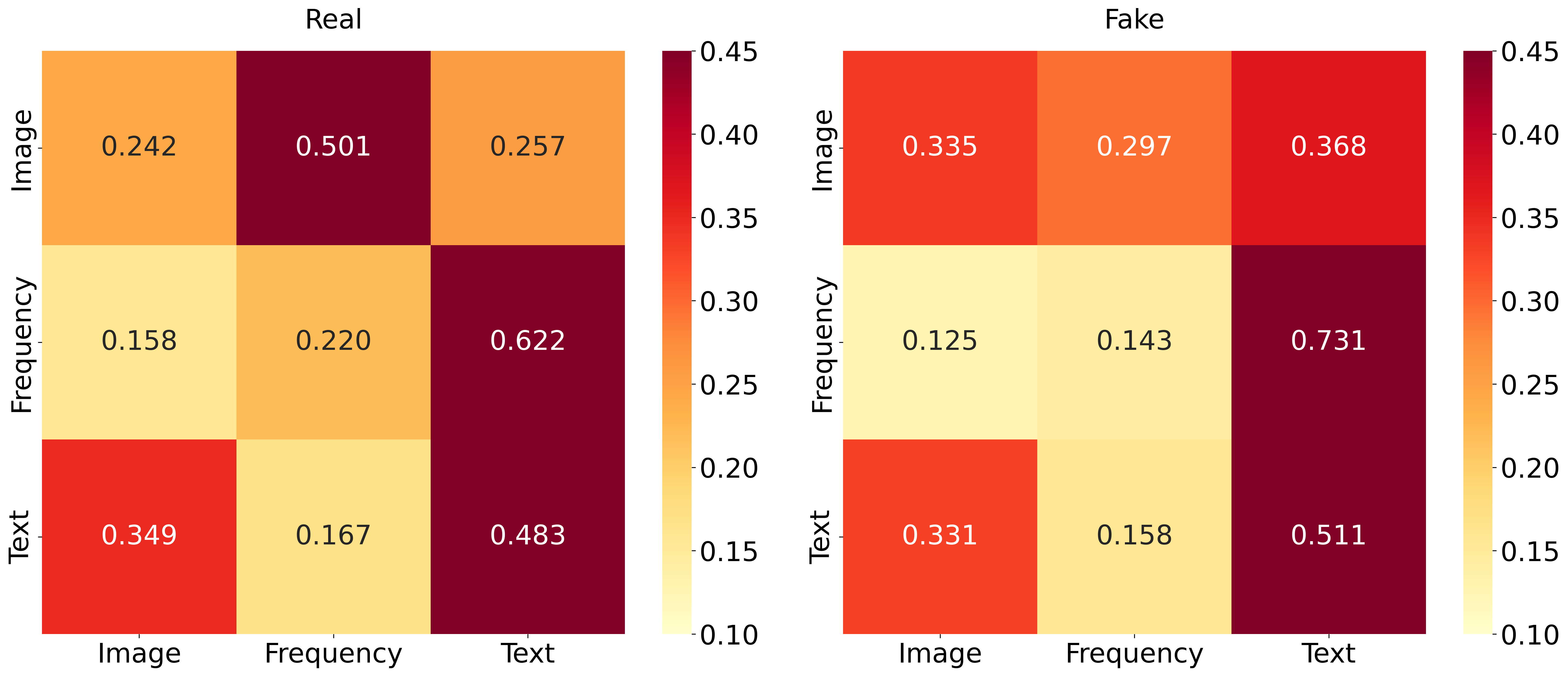}\tabularnewline
    {(a) GLIDE $\scalebox{1.5}{$\rightarrow$}$ GLIDE}\tabularnewline
    \includegraphics[scale=1.0]{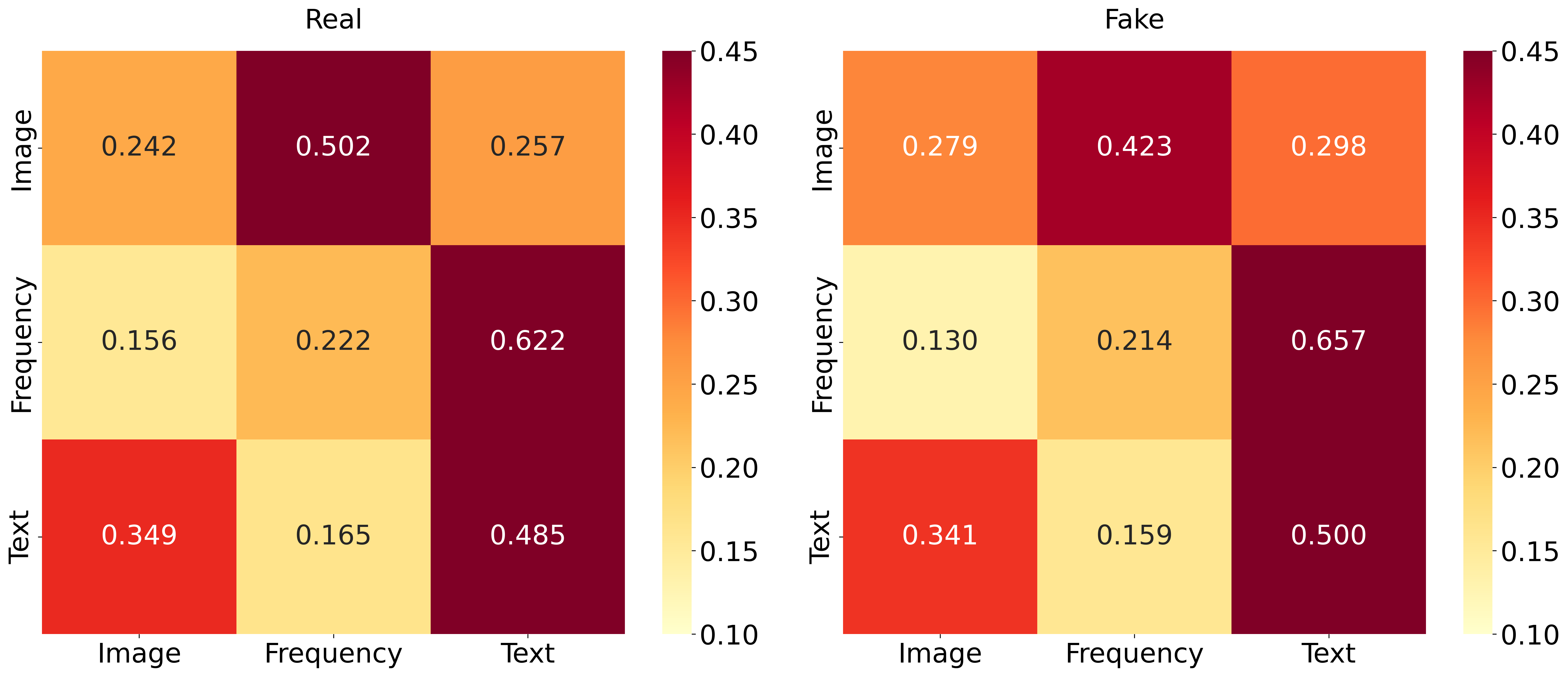}\tabularnewline
    {(b) GLIDE $\scalebox{1.5}{$\rightarrow$}$ ADM}\tabularnewline
    \end{tabular}}\caption{Visualization of attention matrices on the GLIDE and ADM datasets.} \label{fig:attention_map}
\end{figure}

 \begin{center}
        \begin{figure}
        \centering{}\resizebox{0.45\textwidth}{!}{%
        \begin{tabular}{ccccc}
             \includegraphics{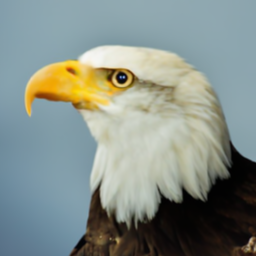} & \includegraphics{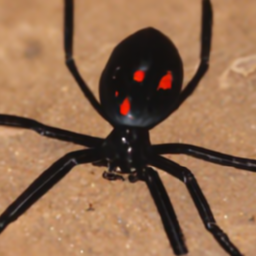} & \includegraphics{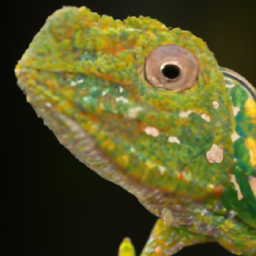} & \includegraphics{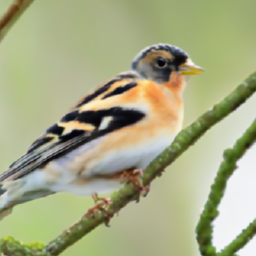} & \includegraphics{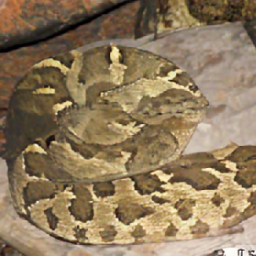}\tabularnewline
            \includegraphics[width=9cm, height=9cm]{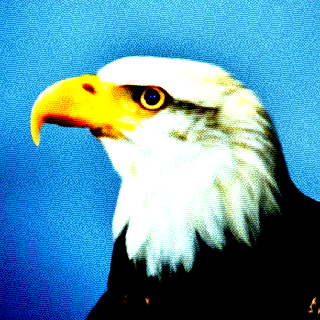} & \includegraphics[width=9cm, height=9cm]{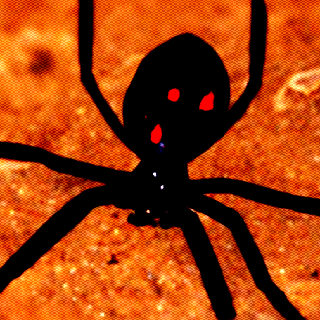} & \includegraphics[width=9cm, height=9cm]{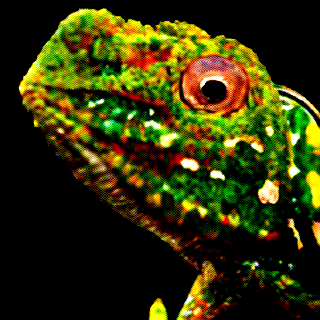} & \includegraphics[width=9cm, height=9cm]{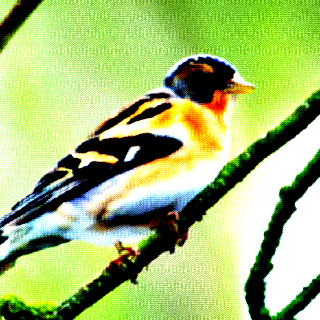} & \includegraphics[width=9cm, height=9cm]{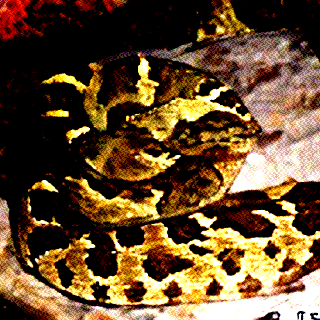}\tabularnewline
        \includegraphics[width=9cm, height=9cm]{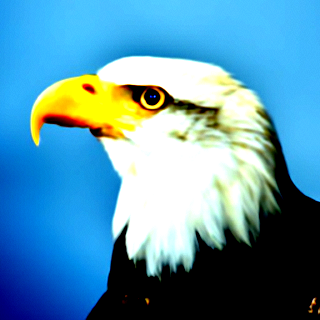} & \includegraphics[width=9cm, height=9cm]{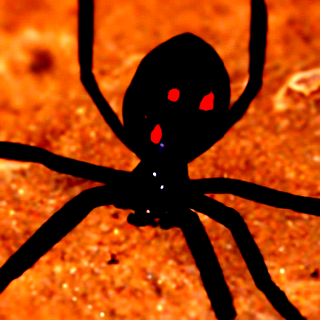} & \includegraphics[width=9cm, height=9cm]{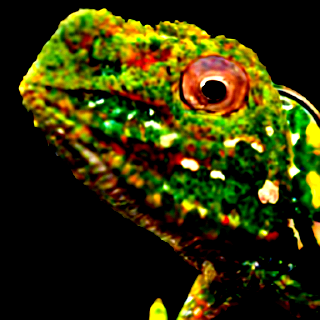} & \includegraphics[width=9cm, height=9cm]{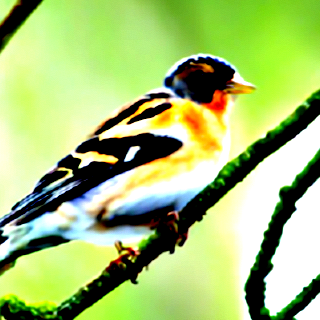} & \includegraphics[width=9cm, height=9cm]{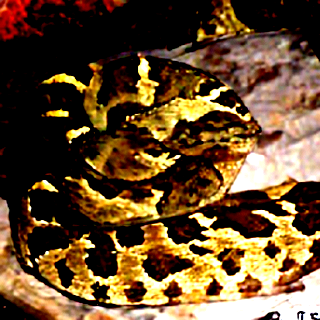}\tabularnewline
        \end{tabular}}\caption{Examples of images from our original fake images (first row), FGSM-generated
        adversarial images (second row), and PGD-generated adversarial images
        (third row).} \label{fig:adversarial_examples}
        \end{figure}
        \par\end{center}

\section{Examples of Adversarial Images}\label{sec:adversarial_examples}
Figure \ref{fig:adversarial_examples} presents examples of our original fake images, adversarial images generated using the FGSM attack, and those produced by the PGD attack.

\begin{table*}[!t]
\begin{center}
\caption{Precision / Recall / F1 / Accuracy scores (\%) for deepfake detection performance on the Natural Scene dataset. Bold blue text indicates intra-domain performance. The highest F1 score is highlighted in bold, and the second-highest is underlined.}
\label{table:natural_scene_full} 
\resizebox{\linewidth}{!}{
\begin{tabular}{|c|c|c|c|c|c|c|c|c|}
\hline 
Methods & \diagbox{Source}{Target} & SD V1.5 & ADM & GLIDE & VQDM & BigGAN & \textbf{Inter-domain Average (IA)} & \textbf{Average of IA} \tabularnewline
\hline 
\multirow{5}{*}{DE-FAKE} & SD V1.5 & \textbf{\textcolor{blue}{98.65}} / \textbf{\textcolor{blue}{98.60}} / \textbf{\textcolor{blue}{98.62}} / \textbf{\textcolor{blue}{98.62}} & 39.53 / 01.00 / 01.96 / 49.73 & 97.13 / 35.82 / 52.34 / 67.38 & 92.21 / 12.53 / 22.06 / 55.74 & 75.86 / 03.88 / 07.39 / 51.32 & 76.18 / 13.31 / 20.94 / 56.04 & \multirow{5}{*}{82.16 / 24.86 / 33.29 / 61.32}\tabularnewline
\cline{2-8}
& ADM & 36.17 / 00.96 / 01.86 / 49.63 & \bluebold{98.01} / \bluebold{99.05} / \bluebold{98.53} / \bluebold{98.52} & 78.18 / 07.59 / 13.83 / 52.74 & 61.54 / 02.35 / 04.53 / 50.44 & 89.78 / 16.53 / 27.92 / 57.32 & 66.42 / 06.86 / 12.04 / 52.53 &\tabularnewline
\cline{2-8}
& GLIDE & 94.99 / 27.70 / 42.89 / 63.12 & 88.58 / 16.96 / 28.47 / 57.39 & \bluebold{98.26} / \bluebold{99.47} / \bluebold{98.86} / \bluebold{98.85} & 91.84 / 21.18 / 34.42 / 59.65 & 97.99 / 91.71 / 94.74 / 94.91 & 93.35 / 39.39 / 50.13 / 68.77 & \tabularnewline
\cline{2-8}
& VQDM & 81.16 / 22.02 / 34.64 / 58.46 & 72.81 / 13.77 / 23.16 / 54.31 & 86.05 / 36.65 / 51.40 / 65.35 & \bluebold{94.62} / \bluebold{96.29} / \bluebold{95.45} / \bluebold{95.41} & 92.33 / 68.00 / 78.32 / 81.18 & 83.09 / 35.11 / 46.88 / 64.83 &\tabularnewline
\cline{2-8}
& BigGAN & 80.43 / 02.08 / 04.05 / 50.79 & 92.63 / 15.60 / 26.71 / 57.18 & 99.15 / 88.82 / 93.70 / 94.03 & 94.88 / 12.00 / 21.31 / 55.68 & \bluebold{99.30} / \bluebold{99.94} / \bluebold{99.62} / \bluebold{99.62} & 91.77 / 29.63 / 36.44 / 64.42 & \tabularnewline
\hline 

\multirow{5}{*}{UnivCLIP} & SD V1.5 & \bluebold{96.47} / \bluebold{98.37} / \bluebold{97.41} / \bluebold{97.39} & 56.14 / 03.78 / 07.09 / 50.41 & 92.88 / 50.65 / 65.55 / 73.38 & 72.25 / 8.12 / 14.60 / 52.50 & 89.35 / 21.71 / 34.93 / 59.56 & 77.66 / 21.07 / 30.54 / 58.96 & \multirow{5}{*}{84.30 / 32.16 / 41.53 / 64.31}\tabularnewline
\cline{2-8}
& ADM & 42.45 / 05.84 / 10.27 / 48.96 & \bluebold{92.73} / \bluebold{92.79} / \bluebold{92.76} / \bluebold{92.76} & 88.74 / 59.35 / 71.13 / 75.91 & 93.32 / 94.47 / 93.89 / 93.85 & 91.98 / 76.88 / 83.76 / 85.09 & 79.12 / 59.14 / 64.76 / 75.95 &\tabularnewline
\cline{2-8}
& GLIDE & 92.77 / 26.69 / 41.45 / 62.30 & 90.91 / 15.96 / 27.15 / 57.18 & \bluebold{98.50} / \bluebold{96.76} / \bluebold{97.63} / \bluebold{97.65} & 91.14 / 16.94 / 28.57 / 57.65 & 95.96 / 40.47 / 56.93 / 69.38 & 92.70 / 25.02 / 38.53 / 61.63 &\tabularnewline
\cline{2-8}
& VQDM & 72.46 / 09.61 / 16.96 / 52.98 & 91.36 / 38.12 / 53.79 / 67.26 & 86.50 / 28.65 / 43.04 / 62.09 & \bluebold{95.90} / \bluebold{98.94} / \bluebold{97.39} / \bluebold{97.35} & 96.78 / 81.24 / 88.33 / 89.26 & 86.78 / 39.41 / 50.53 / 67.90 &\tabularnewline
\cline{2-8}
& BigGAN & 69.70 / 01.29 / 02.54 / 50.37 & 78.90 / 05.08 / 09.55 / 51.86 & 94.87 / 13.06 / 22.96 / 56.18 & 97.50 / 41.29 / 58.02 / 70.12 & \bluebold{99.41} / \bluebold{98.76} / \bluebold{99.09} / \bluebold{99.09} & 85.24 / 15.18 / 23.27 / 57.13 &\tabularnewline
\hline 

\multirow{5}{*}{UnivConv2B} & SD V1.5 & \bluebold{99.89} / \bluebold{99.83} / \bluebold{99.86} / \bluebold{99.86} & 50.00 / 00.06 / 00.12 / 50.00 & 98.32 / 06.88 / 12.86 / 53.38 & 100.0 / 00.76 / 01.52 / 50.38 & 97.56 / 02.35 / 04.60 / 51.15 & 86.47 / 02.51 / 04.78 / 51.23 & \multirow{5}{*}{89.91 / 35.09 / 39.44 / 67.62}\tabularnewline
\cline{2-8}
& ADM & 46.67 / 00.39 / 00.78 / 49.97 & \bluebold{99.53} / \bluebold{99.88} / \bluebold{99.71} / \bluebold{99.70} & 99.23 / 99.00 / 99.12 / 99.12 & 99.11 / 98.71 / 98.91 / 98.91 & 99.47 / 100.0 / 99.74 / 99.74 & 86.12 / 74.53 / 74.64 / 86.94 &\tabularnewline
\cline{2-8}
& GLIDE & 76.92 / 00.56 / 01.12 / 50.20 & 99.58 / 13.89 / 24.38 / 56.91 & \bluebold{99.94} / \bluebold{100.0} / \bluebold{99.97} / \bluebold{99.97} & 100.0 / 19.88 / 33.17 / 59.94 & 100.0 / 54.88 / 70.87 / 77.44 & 94.13 / 22.30 / 32.39 / 61.12 & \tabularnewline
\cline{2-8}
& VQDM & 85.42 / 02.30 / 04.49 / 50.96 & 99.08 / 82.92 / 90.28 / 91.08 & 99.73 / 65.00 / 78.70 / 82.41 & \bluebold{99.65} / \bluebold{99.88} / \bluebold{99.76} / \bluebold{99.76} & 99.43 / 92.00 / 95.57 / 95.74 & 95.92 / 60.56 / 67.26 / 80.05 &\tabularnewline
\cline{2-8}
& BigGAN & 92.31 / 01.35 / 02.66 / 50.62 & 98.95 / 05.56 / 10.52 / 52.75 & 100.0 / 14.06 / 24.65 / 57.03 & 99.77 / 25.06 / 40.06 / 62.50 & \bluebold{99.88} / \bluebold{100.0} / \bluebold{99.94} / \bluebold{99.94} & 97.76 / 11.51 / 19.47 / 55.73 &\tabularnewline
\hline 

\multirow{5}{*}{DCT} & SD V1.5 & \bluebold{95.86} / \bluebold{97.64} / \bluebold{96.74} / \bluebold{96.71} & 49.02 / 44.27 / 46.52 / 49.11 & 73.40 / 55.35 / 63.11 / 67.65 & 90.24 / 90.29 / 90.27 / 90.26 & 56.07 / 50.00 / 52.86 / 55.41 & 67.18 / 59.98 / 63.19 / 65.61 & \multirow{5}{*}{64.19 / 61.61 / 62.72 / 63.26}\tabularnewline
\cline{2-8}
& ADM & 63.69 / 83.26 / 72.17 / 67.89 & \bluebold{99.41} / \bluebold{99.00} / \bluebold{99.20} / \bluebold{99.20} & 56.60 / 72.12 / 63.42 / 58.41 & 65.81 / 81.06 / 72.64 / 69.47 & 47.07 / 56.29 / 51.27 / 46.50 & 58.29 / 73.18 / 64.88 / 60.57 &\tabularnewline
\cline{2-8}
& GLIDE & 56.23 / 46.12 / 50.68 / 55.11 & 58.77 / 45.15 / 51.07 / 56.74 & \bluebold{99.35} / \bluebold{99.06} / \bluebold{99.20} / \bluebold{99.21} & 51.98 / 41.00 / 45.84 / 51.56 & 77.09 / 67.88 / 72.19 / 73.85 & 61.02 / 50.04 / 54.95 / 59.32 &\tabularnewline
\cline{2-8}
& VQDM & 87.10 / 84.94 / 86.01 / 86.18 & 78.23 / 69.68 / 73.71 / 75.15 & 78.30 / 54.76 / 64.45 / 69.79 & \bluebold{96.37} / \bluebold{98.29} / \bluebold{97.32} / \bluebold{97.29} & 77.13 / 68.82 / 72.74 / 74.21 & 80.19 / 69.55 / 74.23 / 76.33 &\tabularnewline
\cline{2-8}
& BigGAN & 58.06 / 56.07 / 57.04 / 57.78 & 66.60 / 59.87 / 63.06 / 64.92 & 95.01 / 98.53 / 96.74 / 96.68 & 60.41 / 54.29 / 57.19 / 59.35 & \bluebold{100.0} / \bluebold{100.0} / \bluebold{100.0} / \bluebold{100.0} & 70.02 / 67.19 / 68.51 / 69.68 &\tabularnewline
\hline 

\multirow{5}{*}{CAMME} & SD V1.5 & \bluebold{99.61} / \bluebold{99.38} / \bluebold{99.49} / \bluebold{99.49} & 50.06 / 98.88 / 66.47 / 50.12 & 59.88 / 99.47 / 74.76 / 66.41 & 52.03 / 99.59 / 68.35 / 53.88 & 56.51 / 99.29 / 72.03 / 61.44 & 54.62 / 99.31 / 70.40 / 57.96 & \multirow{5}{*}{68.68 / 99.35 / \textbf{79.94} / 70.71}\tabularnewline
\cline{2-8}
& ADM & 49.73 / 98.93 / 66.19 / 49.47 & \bluebold{99.88} / \bluebold{99.05} / \bluebold{99.47} / \bluebold{99.47} & 92.52 / 99.00 / 95.65 / 95.50 & 99.12 / 98.82 / 98.97 / 98.97 & 99.47 / 99.06 / 99.26 / 99.26 & 85.21 / 98.95 / 90.02 / 85.80 &\tabularnewline
\cline{2-8}
& GLIDE & 50.33 / 99.94 / 66.94 / 50.65 & 53.13 / 99.82 / 69.35 / 55.88 & \bluebold{99.59} / \bluebold{99.71} / \bluebold{99.65} / \bluebold{99.65} & 58.09 / 99.71 / 73.41 / 63.88 & 58.94 / 99.82 / 74.12 / 65.15 & 55.12 / 99.82 / 70.96 / 58.89 & \tabularnewline
\cline{2-8}
& VQDM & 50.13 / 99.78 / 66.73 / 50.25 & 89.37 / 99.41 / 94.12 / 93.79 & 73.38 / 99.88 / 84.60 / 81.82 & \bluebold{99.94} / \bluebold{99.65} / \bluebold{99.79} / \bluebold{99.79} & 86.39 / 99.29 / 92.39 / 91.82 & 74.82 / 99.59 / 84.46 / 79.42 &\tabularnewline
\cline{2-8}
& BigGAN & 50.54 / 99.72 / 67.08 / 51.07 & 54.57 / 99.47 / 70.48 / 58.33 & 60.05 / 99.65 / 74.94 / 66.68 & 64.73 / 99.41 / 78.40 / 72.62 & \bluebold{100.0} / \bluebold{99.53} / \bluebold{99.76} / \bluebold{99.76} & 57.47 / 99.56 / 72.73 / 62.18 &\tabularnewline
\hline 
\end{tabular}
}
\end{center}
\end{table*}

\begin{table*}[!t]
    \caption{Precision / Recall / F1 / Accuracy scores (\%) for deepfake detection performance on the Face dataset. Bold blue text indicates intra-domain performance. The highest F1 score is highlighted in bold, and the second-highest is underlined.} 
    \label{table:face_full} 
    \begin{center}
    \resizebox{\linewidth}{!}{
    \begin{tabular}{|c|c|c|c|c|c|c|c|}
    \hline 
    \textbf{Methods} & \diagbox{Source}{Target} & CelPG & CelStar & CelGlow & YouFace & \textbf{Inter-domain Average (IA)} & \textbf{Average of IA}\tabularnewline
    \hline 
    \multirow{4}{*}{DE-FAKE} & CelPG & \bluebold{99.36} / \bluebold{92.70} / \bluebold{95.91} / \bluebold{96.05} & 96.91 / 47.00 / 63.30 / 72.75 & 71.15 / 03.70 / 07.03 / 51.10 & 100.0 / 00.80 / 01.59 / 50.40 & 89.35 / 17.17 / 23.97 / 58.08 & \multirow{4}{*}{57.65 / 13.65 / 15.98 / 52.16}\tabularnewline
    \cline{2-7}
    & CelStar & 100.0 / 01.20 / 02.37 / 50.60 & \bluebold{99.80} / \bluebold{99.90} / \bluebold{99.85} / \bluebold{99.85} & 93.55 / 02.90 / 05.63 / 51.35 & 00.00 / 00.00 / 00.00 / 49.95 & 64.52 / 01.37 / 02.67 / 50.63 &\tabularnewline
    \cline{2-7}
    & CelGlow & 00.00 / 00.00 / 00.00 / 50.00 & 100.0 / 00.20 / 00.40 / 50.10 & \bluebold{100.0} / \bluebold{98.50} / \bluebold{99.24} / \bluebold{99.25} & 05.71 / 00.60 / 01.09 / 45.35 & 35.24 / 00.27 / 00.50 / 48.48 &\tabularnewline
    \cline{2-7}
    & YouFace & 61.28 / 73.90 / 67.00 / 63.60 & 54.39 / 31.00 / 39.49 / 52.50 & 08.77 / 02.50 / 03.89 / 38.25 & \bluebold{71.97} / \bluebold{91.90} / \bluebold{80.72} / \bluebold{78.05} & 41.48 / 35.80 / 36.79 / 51.45 &\tabularnewline
    \hline 
    \multirow{4}{*}{UnivCLIP} & CelPG & \bluebold{100.0} / \bluebold{99.20} / \bluebold{99.60} / \bluebold{99.60} & 95.80 / 89.00 / 92.28 / 92.55 & 83.40 / 19.60 / 31.74 / 57.85 & 99.52 / 41.14 / 58.21 / 70.41 & 92.91 / 49.91 / 60.74 / 73.60 & \multirow{4}{*}{88.74 / 44.94 / \uline{53.21} / 69.87}\tabularnewline
    \cline{2-7}
    & CelStar & 100.0 / 62.64 / 77.03 / 81.31 & \bluebold{99.70} / \bluebold{98.80} / \bluebold{99.25} / \bluebold{99.25} & 91.18 / 03.10 / 06.00 / 51.40 & 99.61 / 25.60 / 40.73 / 62.67 & 96.93 / 30.45 / 41.25 / 65.13 &\tabularnewline
    \cline{2-7}
    & CelGlow & 89.74 / 03.50 / 06.73 / 51.52 & 94.48 / 58.20 / 72.03 / 77.40 & \bluebold{96.70} / \bluebold{99.50} / \bluebold{98.08} / \bluebold{98.05} & 68.21 / 20.52 / 31.55 / 55.39 & 84.14 / 27.41 / 36.77 / 61.44 &\tabularnewline
    \cline{2-7}
    & YouFace & 89.98 / 96.90 / 93.31 / 93.05 & 85.84 / 89.10 / 87.44 / 87.20 & 67.11 / 30.00 / 41.47 / 57.65 & \bluebold{86.88} / \bluebold{99.60} / \bluebold{92.81} / \bluebold{92.27} & 80.98 / 72.00 / 74.07 / 79.30 &\tabularnewline
    \hline 
    \multirow{4}{*}{UnivConv2B} & CelPG & \bluebold{100.0} / \bluebold{99.80} / \bluebold{99.90} / \bluebold{99.90} & 78.99 / 78.60 / 78.80 / 78.85 & 30.56 / 09.20 / 14.14 / 44.15 & 100.0 / 18.03 / 30.55 / 58.93 & 69.85 / 35.28 / 41.16 / 60.64 & \multirow{4}{*}{45.56 / 26.08 / 28.94 / 51.41}\tabularnewline
    \cline{2-7}
    & CelStar & 00.00 / 00.00 / 00.00 / 49.98 & \bluebold{100.0} / \bluebold{99.90} / \bluebold{99.95} / \bluebold{99.95} & 00.00 / 00.00 / 00.00 / 50.00 & 00.00 / 00.00 / 00.00 / 00.50 & 00.00 / 00.00 / 00.00 / 33.49 &\tabularnewline
    \cline{2-7}
    & CelGlow & 57.14 / 00.40 / 00.79 / 50.02 & 100.0 / 50.70 / 67.29 / 75.35 & \bluebold{100.0} / \bluebold{100.0} / \bluebold{100.0} / \bluebold{100.0} & 47.17 / 27.39 / 34.66 / 48.25 & 68.10 / 26.16 / 34.25 / 57.87 &\tabularnewline
    \cline{2-7}
    & YouFace & 67.21 / 98.50 / 79.90 / 75.21 & 24.62 / 09.60 / 13.81 / 40.10 & 41.08 / 20.50 / 27.35 / 45.55 & \bluebold{97.64} / \bluebold{98.90} / \bluebold{98.27} / \bluebold{98.25} & 44.30 / 42.87 / 40.35 / 53.62 &\tabularnewline
    \hline 
    \multirow{4}{*}{DCT} & CelPG & \bluebold{94.09} / \bluebold{90.80} / \bluebold{92.42} / \bluebold{92.55} & 10.48 / 08.10 / 09.14 / 19.45 & 40.71 / 42.30 / 41.49 / 40.35 & 48.23 / 31.40 / 38.04 / 48.85 & 33.14 / 27.27 / 29.56 / 36.22 & \multirow{4}{*}{45.48 / 45.68 / 44.65 / 46.21}\tabularnewline
    \cline{2-7}
    & CelStar & 34.03 / 35.70 / 34.85 / 33.25 & \bluebold{99.90} / \bluebold{100.0} / \bluebold{99.95} / \bluebold{99.95} & 73.50 / 43.00 / 54.26 / 63.75 & 50.29 / 68.30 / 57.93 / 50.40 & 52.61 / 49.00 / 49.01 / 49.13 &\tabularnewline
    \cline{2-7}
    & CelGlow & 44.28 / 44.90 / 44.59 / 44.20 & 24.71 / 17.10 / 20.21 / 32.50 & \bluebold{99.20} / \bluebold{99.50} / \bluebold{99.35} / \bluebold{99.35} & 48.42 / 59.60 / 53.43 / 48.05 & 39.14 / 40.53 / 39.41 / 41.58 &\tabularnewline
    \cline{2-7}
    & YouFace & 47.68 / 66.90 / 55.68 / 46.75 & 69.41 / 84.20 / 76.10 / 73.55 & 53.93 / 46.70 / 50.05 / 53.40 & \bluebold{80.59} / \bluebold{71.40} / \bluebold{75.72} / \bluebold{77.10} & 57.01 / 65.93 / 60.61 / 57.90 &\tabularnewline
    \hline 
    \multirow{4}{*}{CAMME} & CelPG & \bluebold{99.90} / \bluebold{99.60} / \bluebold{99.75} / \bluebold{99.75} & 65.06 / 89.00 / 75.17 / 70.60 & 47.85 / 89.00 / 62.24 / 46.00 & 54.64 / 100.0 / 70.67 / 58.50 & 55.85 / 92.67 / 69.36 / 58.37 & \multirow{4}{*}{56.15 / 86.88 / \textbf{66.46} / 55.96}\tabularnewline
    \cline{2-7}
    & CelStar & 50.03 / 100.0 / 66.69 / 50.05 & \bluebold{100.0} / \bluebold{100.0} / \bluebold{100.0} / \bluebold{100.0} & 50.08 / 100.0 / 66.73 / 50.15 & 49.97 / 99.80 / 66.60 / 49.95 & 50.03 / 99.93 / 66.67 / 50.05 &\tabularnewline
    \cline{2-7}
    & CelGlow & 50.13 / 100.0 / 66.78 / 50.25 & 63.11 / 99.90 / 77.35 / 70.75 & \bluebold{100.0} / \bluebold{99.90} / \bluebold{99.95} / \bluebold{99.95} & 52.82 / 59.00 / 55.74 / 53.15 & 55.35 / 86.30 / 66.62 / 58.05 &\tabularnewline
    \cline{2-7}
    & YouFace & 94.63 / 61.70 / 74.70 / 79.10 & 45.87 / 72.10 / 56.07 / 43.50 & 49.62 / 72.10 / 58.79 / 49.45 & \bluebold{97.97} / \bluebold{91.70} / \bluebold{94.73} / \bluebold{94.90} & 63.37 / 68.63 / 63.19 / 57.35 &\tabularnewline
    \hline 
    \end{tabular}
    }
    \end{center}
\end{table*}

\section{Attention Map Visualization}
To provide insights into the interactions among the embedding types (\textit{Image}, \textit{Text}, and \textit{Frequency}), we visualize the attention matrices of CAMME in Figure \ref{fig:attention_map}. Specifically, we display the average attention weights across test images from both the GLIDE and ADM datasets. These attention maps reveal distinct interaction patterns for each domain, indicating that the model effectively learns to capture relationships between different embeddings while adapting to the unique characteristics of each dataset. The key findings from the attention maps are summarized below:
\begin{itemize}
    \item \textbf{Consistency in Frequency Cross-Attention}: Across both real and fake images, Frequency embeddings demonstrate strong self-attention (0.500 in all cases), suggesting that the Frequency component retains its unique characteristics independent of the real/fake distinction or model type (GLIDE or ADM).

    \item \textbf{Variation in Image-to-Frequency Attention}: Image-to-Frequency attention (row 1, column 2) shows significant differences between real and fake images. For instance, the attention weight is 0.501 for real images in the GLIDE $\rightarrow$ GLIDE configuration, whereas it decreases to 0.297 for fake images. This indicates the model’s reliance on frequency-based details and textures that are characteristic of real images, which typically exhibit more complex and diverse frequency components. Conversely, for fake images, frequency information may be less informative or even misleading due to artifacts and inconsistencies, resulting in a lower Image-to-Frequency attention score when analyzing synthetic data.

    \item \textbf{Role of Text Embeddings in Fake Images}: Text-to-Image attention (row 3, column 1) and Text Self-Attention (row 3, column 3) scores are consistently higher, highlighting the importance of text embeddings in capturing semantic information and guiding the model to focus on relevant visual features. Fake images from diffusion-based models, for instance, exhibit high Text Self-Attention scores (0.500), suggesting that these images may contain text-related artifacts or repetitive patterns that reinforce certain features within the Text embedding. This phenomenon could serve as a potential indicator for detecting Diffusion-generated fake images.

    \item \textbf{Transferable Attention Map}: The similarity observed between attention matrices for different configurations (e.g., fake class in GLIDE $\rightarrow$ GLIDE and GLIDE $\rightarrow$ ADM) indicates a strong transferability of the model's attention mechanism from the GLIDE to the ADM dataset. The model appears to rely on stable attention patterns, particularly in self-attention for Frequency and Text embeddings, which remain effective across datasets. Minor variations, such as a slightly higher Text-to-Text attention score in Stable diffusion-based fake images, suggest that the model can subtly adapt to different artifact types, enhancing its potential for generalization across generative models.
\end{itemize}

\section{DCT spectra}

\begin{figure}[t!]
    \centering{}\resizebox{0.5\textwidth}{!}{%
    \begin{tabular}{>{\raggedright}cccccc}
        \multirow{-25}{*}{\centering \Huge Real} & \includegraphics{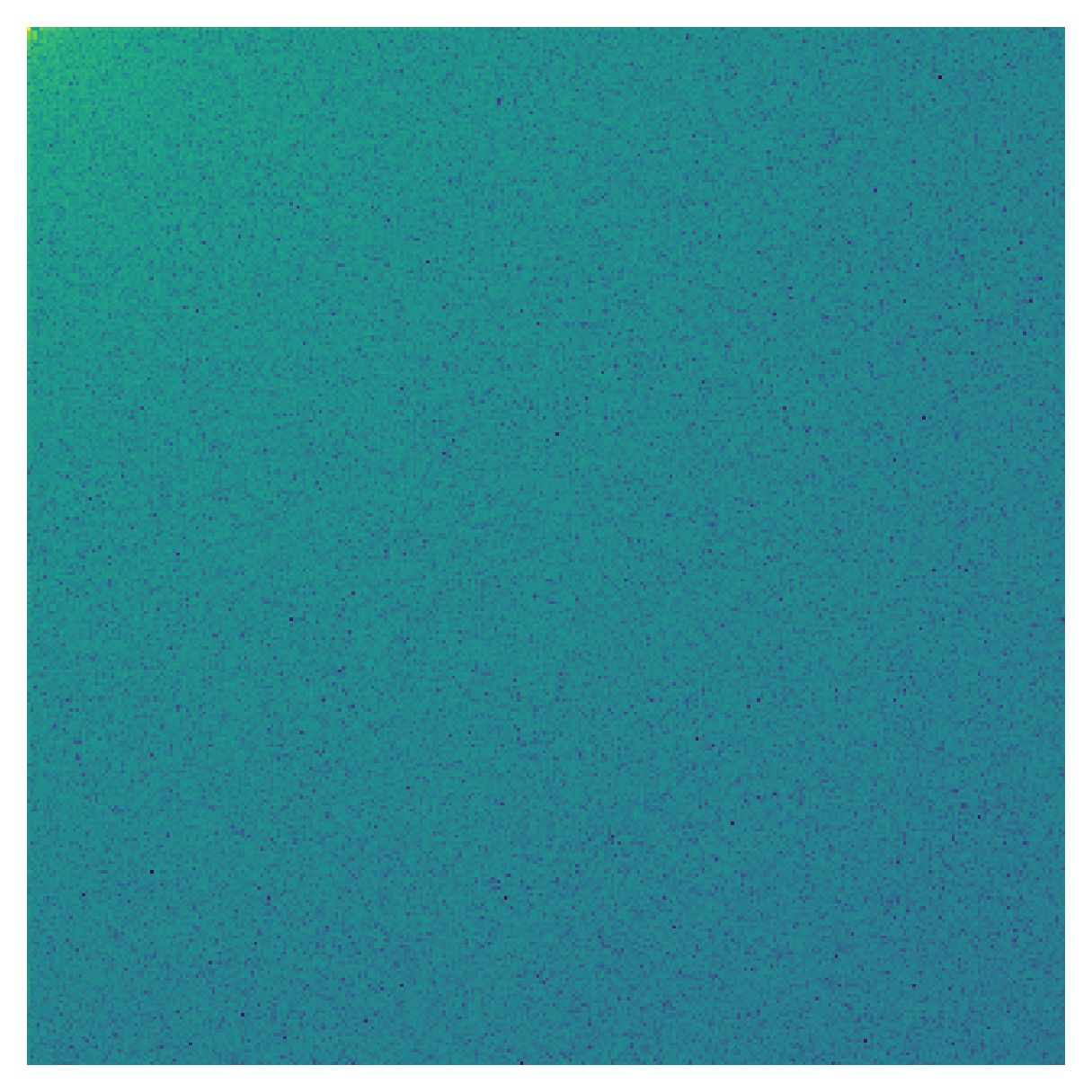} & \includegraphics{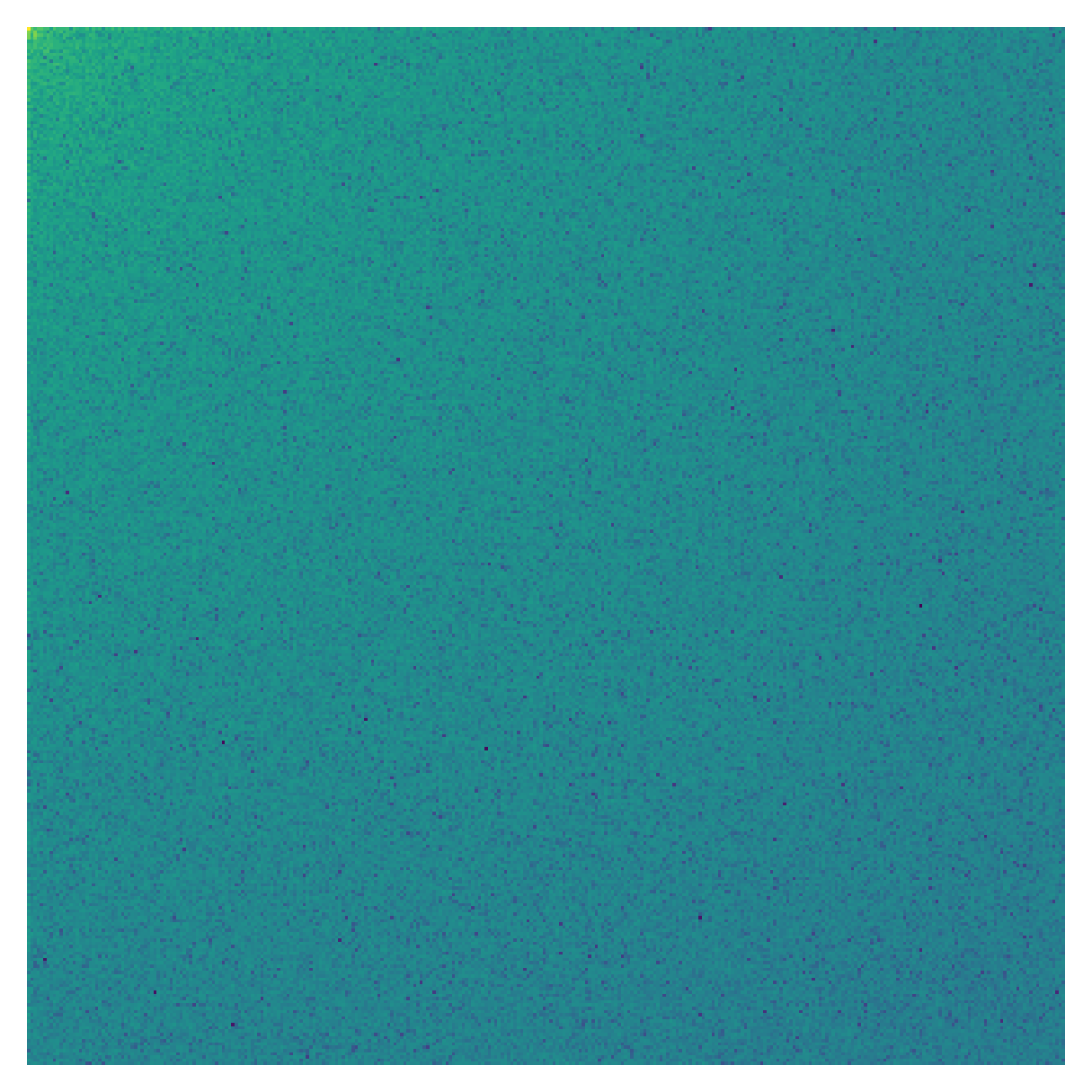} & \includegraphics{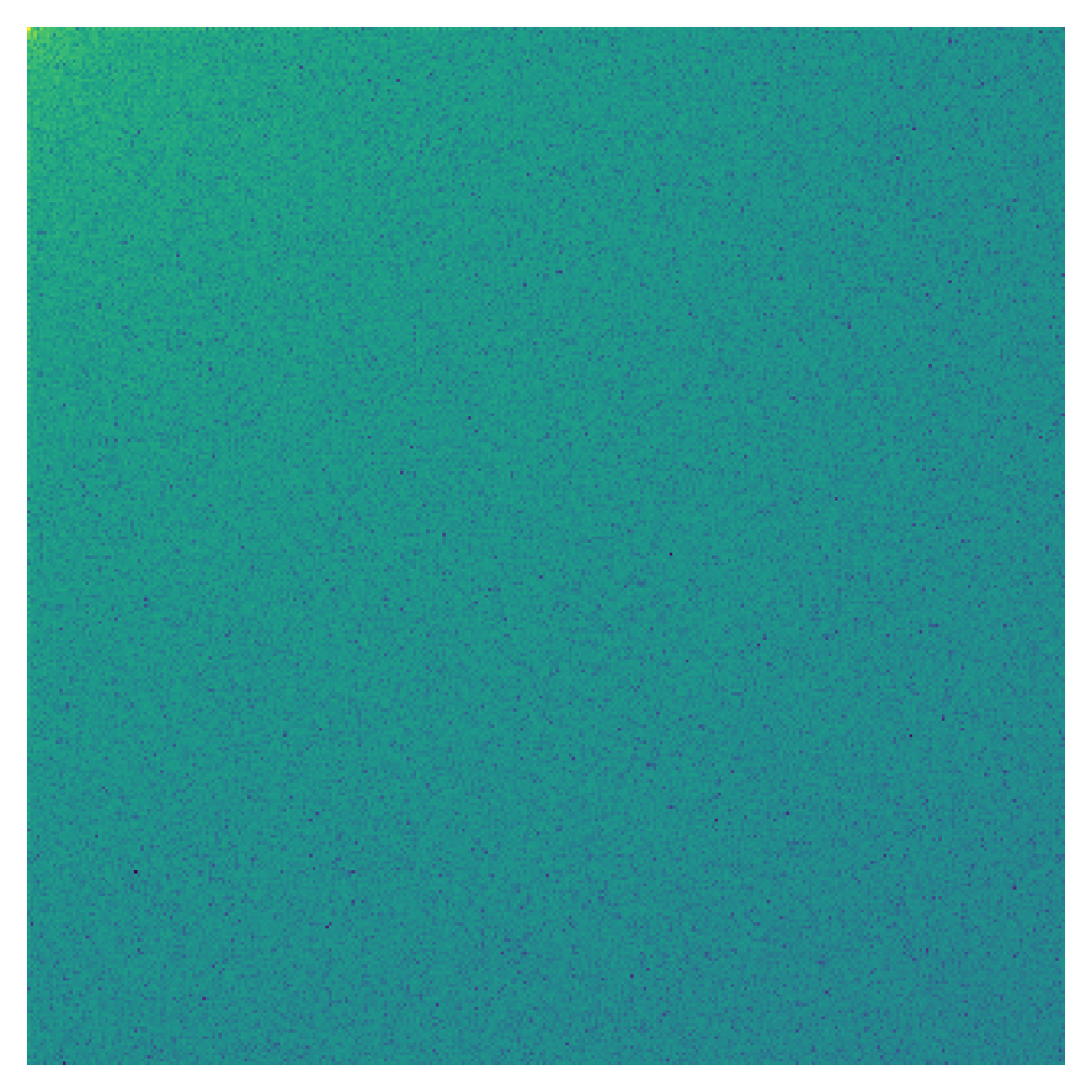} & \includegraphics{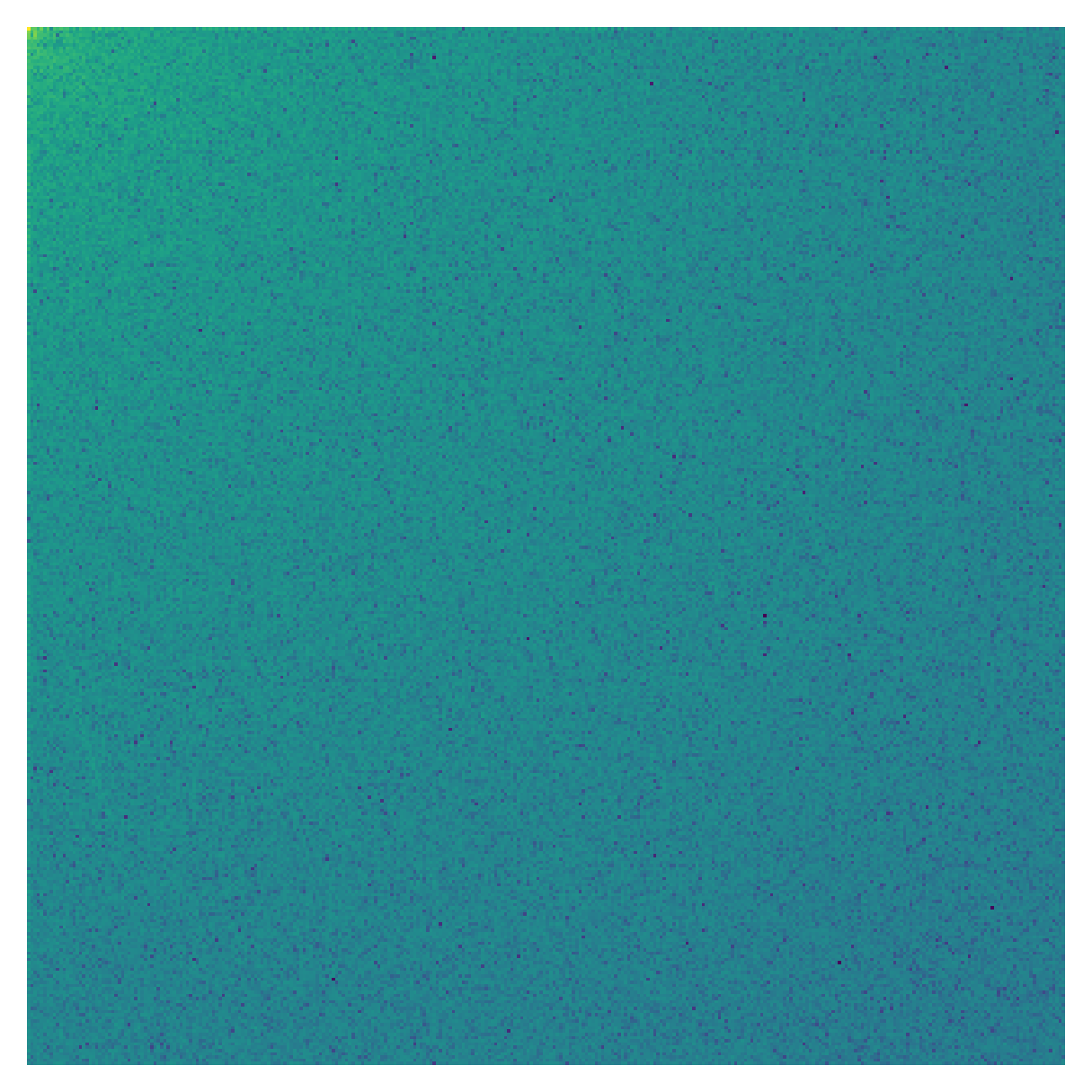} & \includegraphics{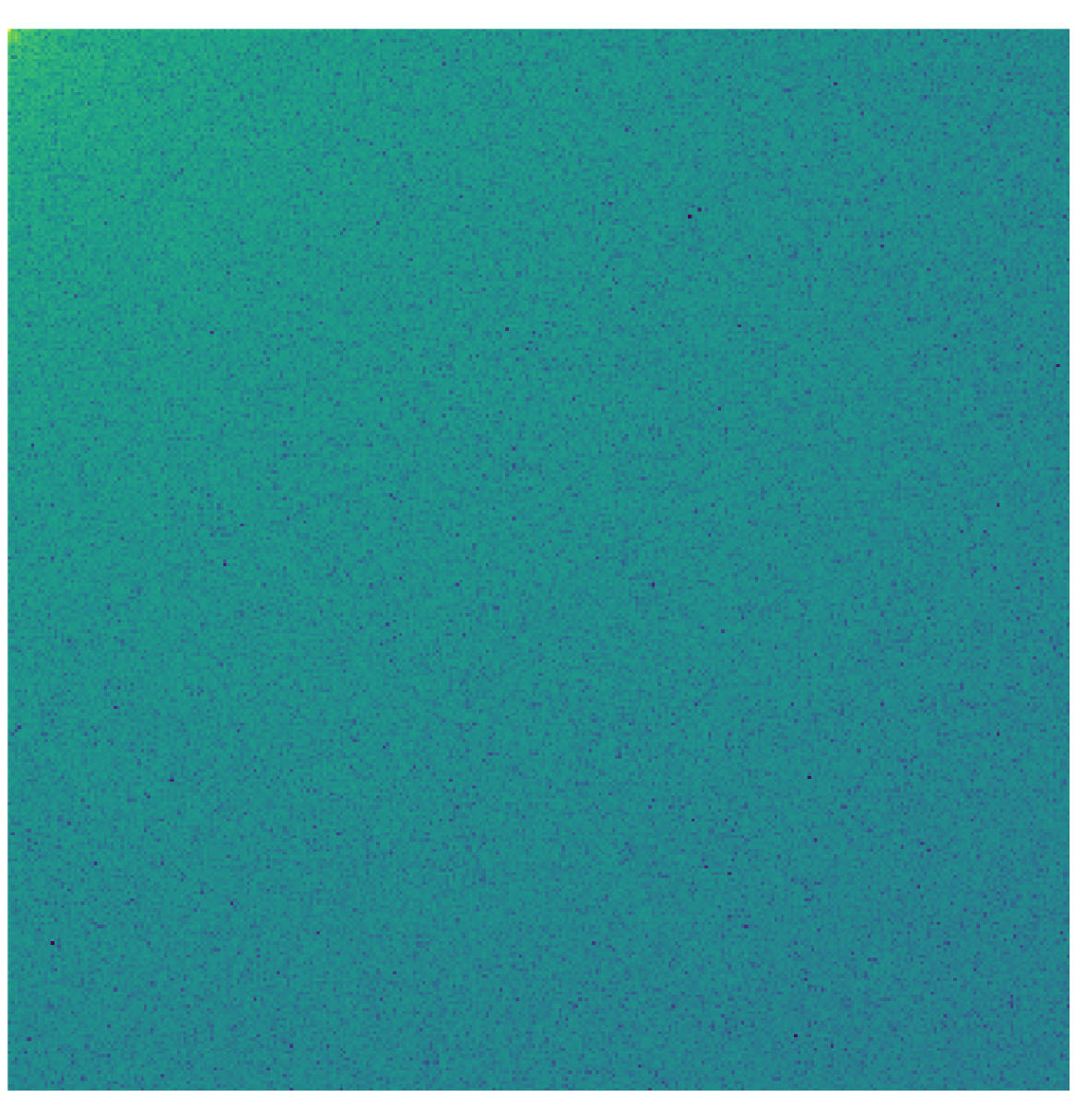}\tabularnewline
        \multirow{-25}{*}{\centering \Huge Fake} & \includegraphics{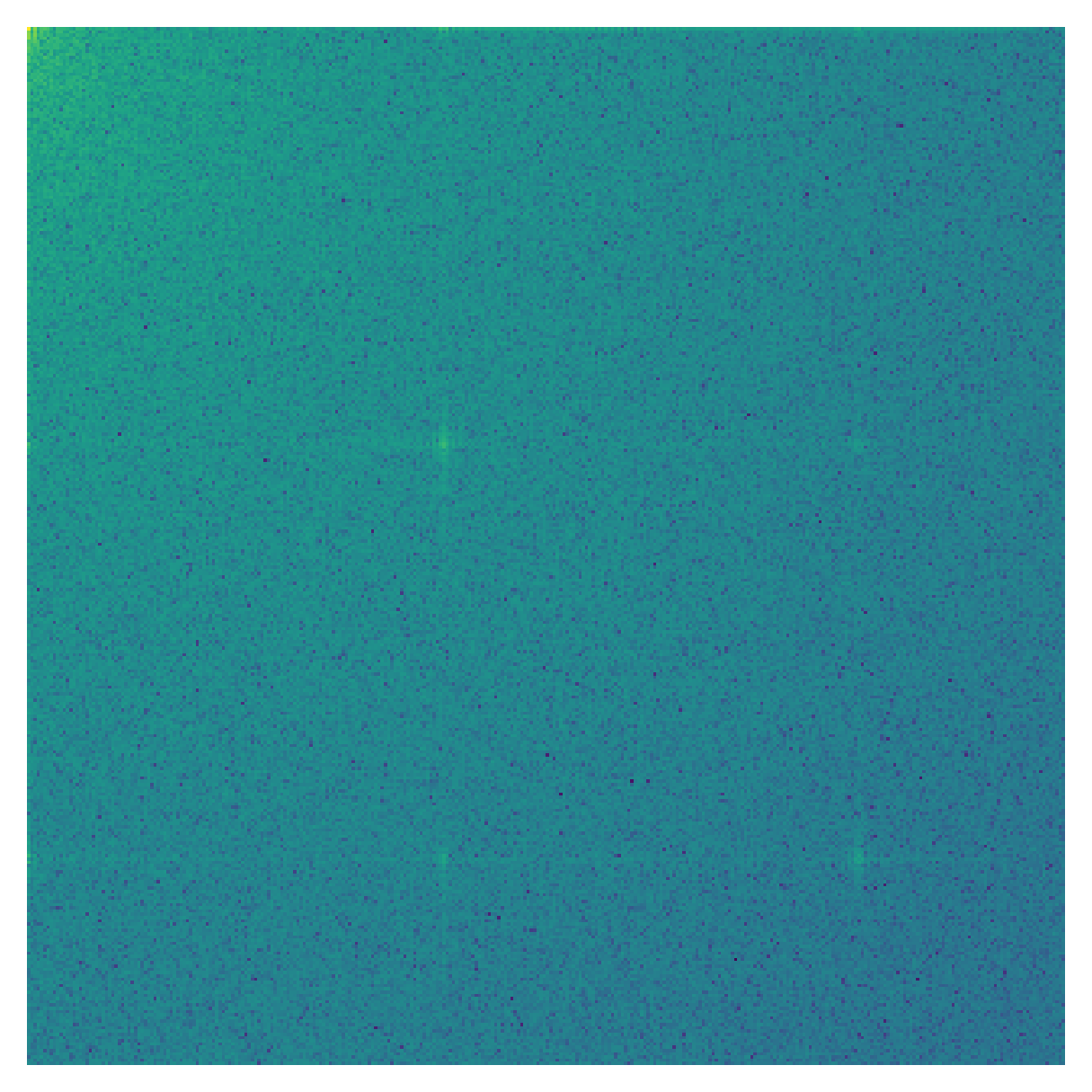} & \includegraphics{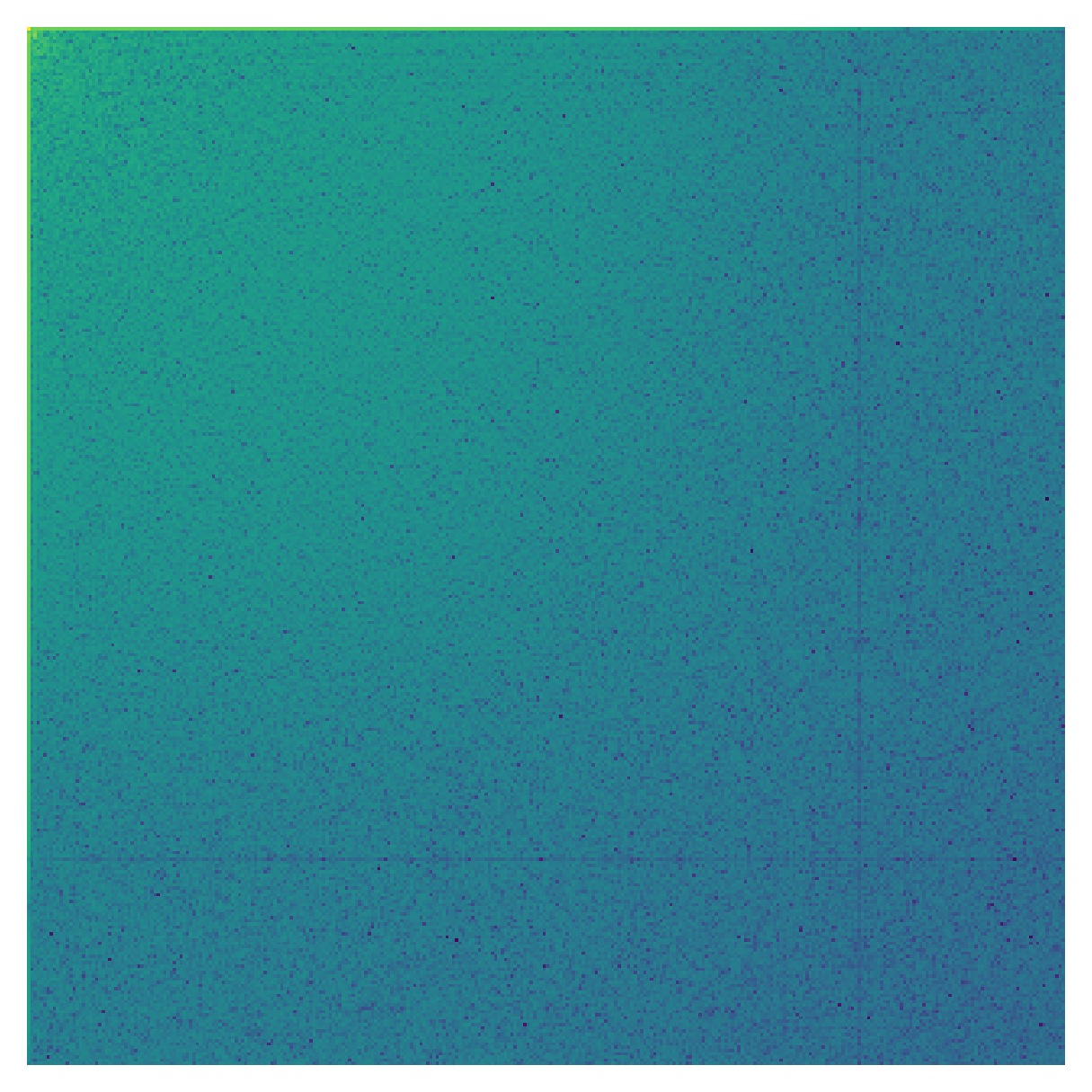} & \includegraphics{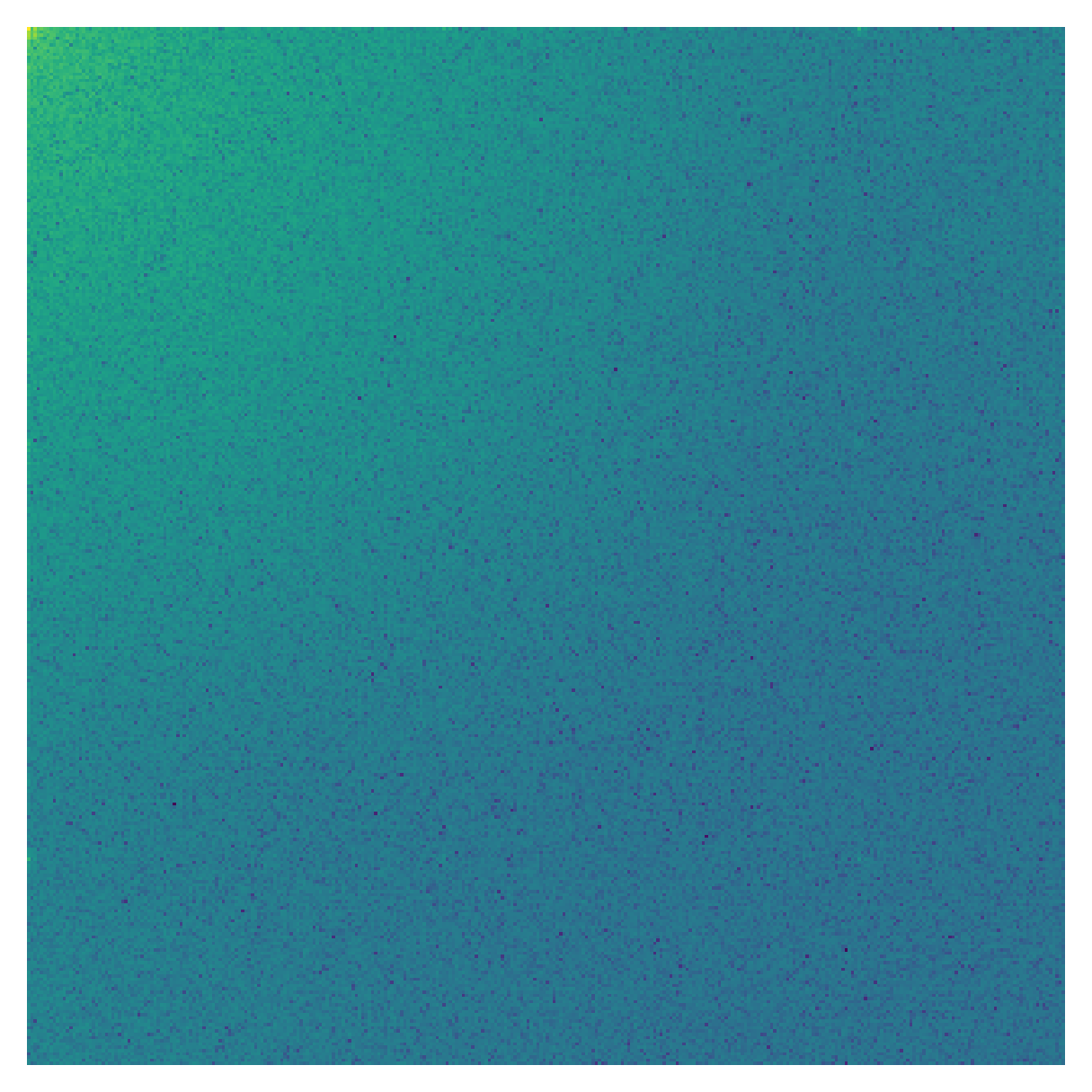} & \includegraphics{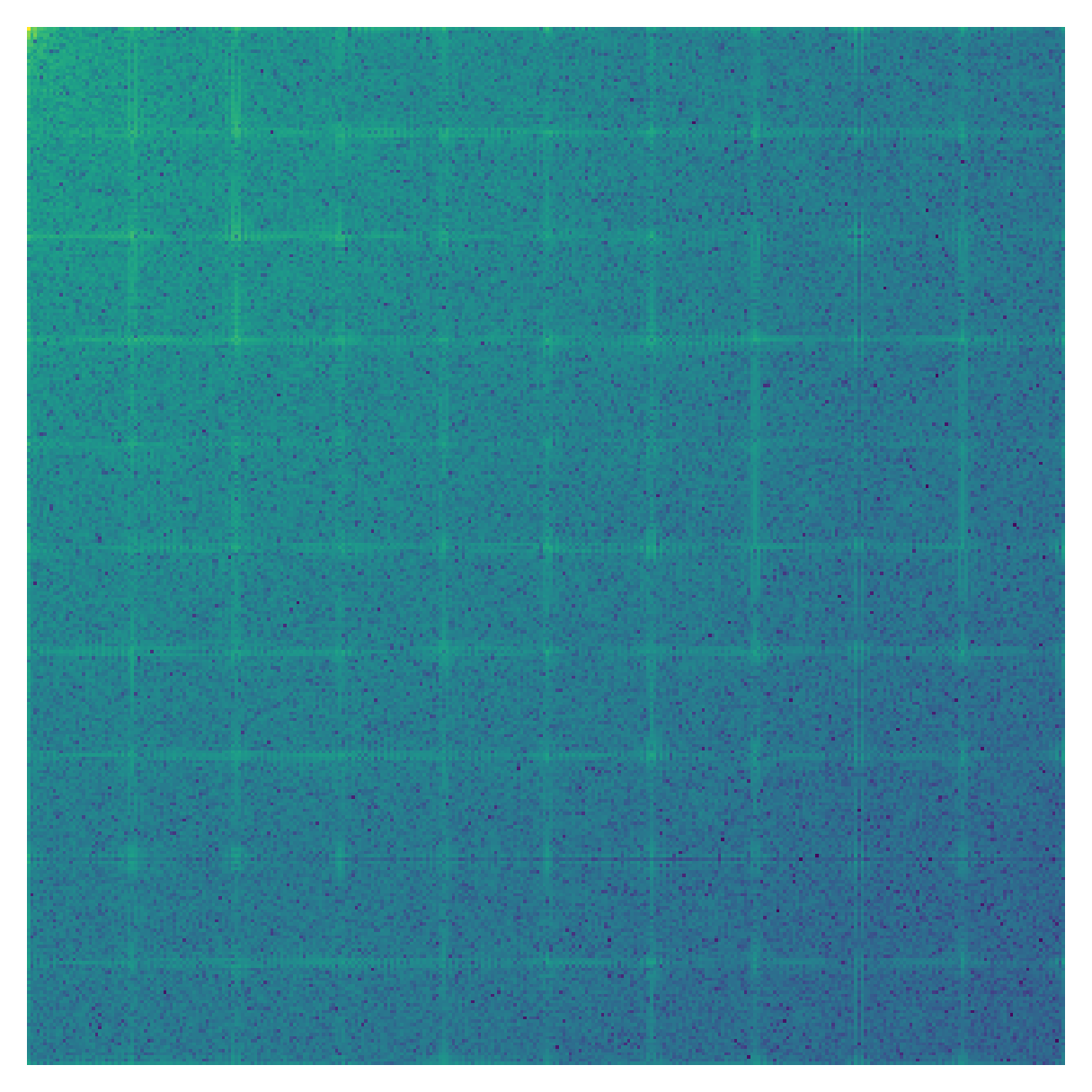} & \includegraphics{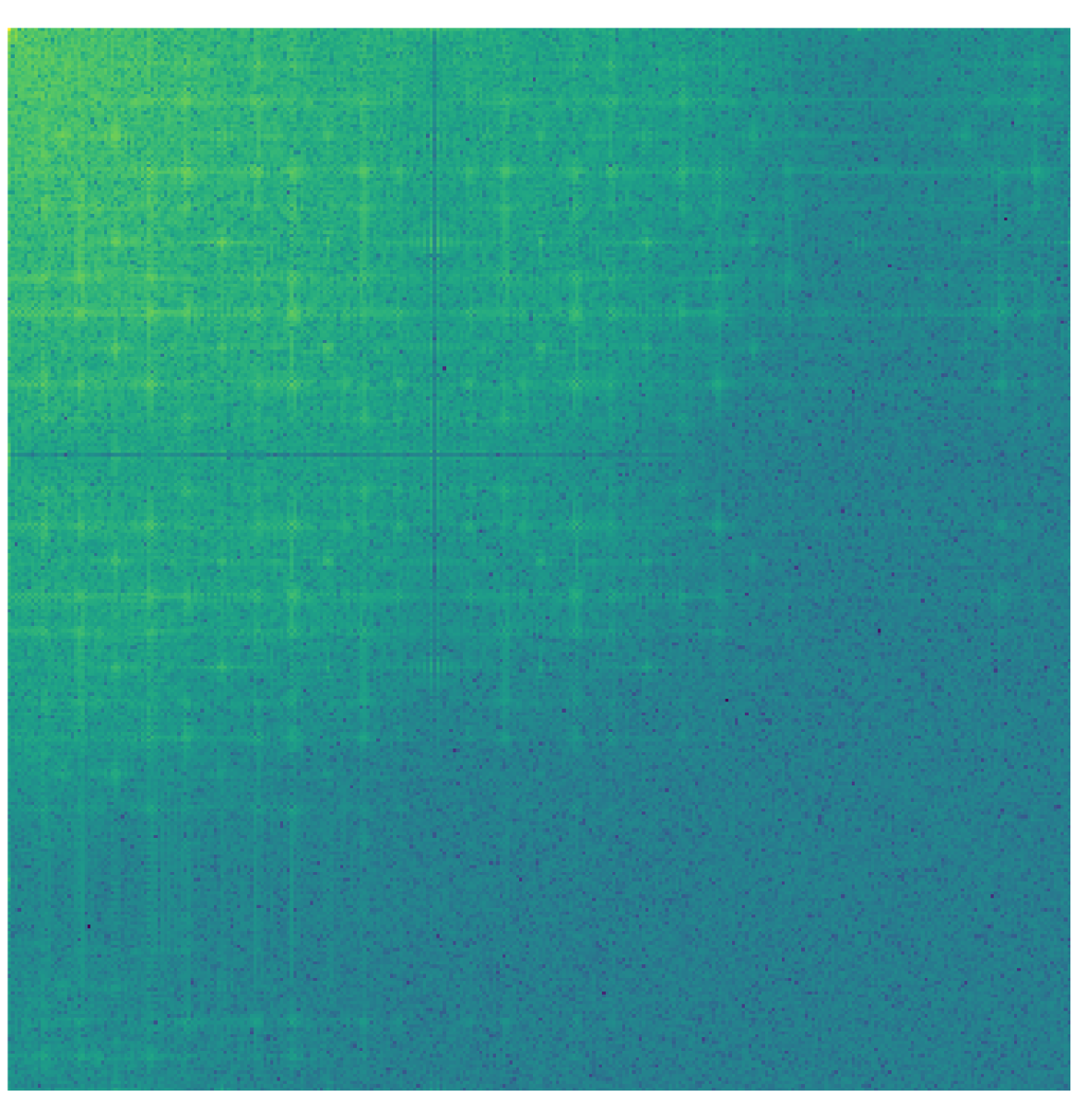}\tabularnewline
     & {\Huge(a) SD V1.5} & {\Huge(b) ADM} & {\Huge(c) GLIDE} & {\Huge(d) VQDM} & {\Huge(e) BigGAN}\tabularnewline
    \end{tabular}}\caption{Average DCT spectrum of real (top row) and fake (bottom row) images generated by various models. Values are shown on a log scale and limited to the range $[-25, 5]$ for improved visualization.} \label{fig:dct_spectra}
\end{figure}

In this section, we analyze the DCT spectra of real and fake images. Following \cite{rickerDetectionDiffusionModel2024}, grayscale-converted test images are transformed using the Discrete Cosine Transform (DCT, (\ref{eq:dct})) to extract frequency components. The average DCT spectra are computed across 1,700 images per class (see Table \ref{table:dataset_details}) and visualized after applying a logarithmic transformation. The results are shown in Figure \ref{fig:dct_spectra}.

Our observations indicate that diffusion-based models trained with classifier-free guidance (e.g., ADM, GLIDE) produce DCT spectra with smooth, continuous energy distributions, with high energy concentrated in the lower frequency bands (top-left corner), closely resembling the characteristics of real images. We attribute this to the influence of prompt conditioning, which enables these models to generate detailed, context-specific, and semantically rich images. This conditioning provides fine-grained control over the generated content and significantly reduces artifacts, resulting in more natural-looking images.

In contrast, the DCT spectra of GAN-generated images display a more fragmented energy distribution, with a noticeable spread into higher frequency bands. This distribution is influenced by factors such as single-step generation, mode collapse, and instability during GAN training. Additionally, the DCT spectra of VQDM (Vector Quantized Diffusion Model) reveal a grid-like pattern due to its use of discrete codebook embeddings to represent synthetic images. Through quantization, VQDM converts the continuous pixel space into discrete latent codes, which are then processed in the denoising steps, often resulting in regular, repeating high-frequency components. This spectral pattern reflects unnatural periodicity, frequently manifesting in the spatial domain as visible blockiness or repetitive artifacts.

\end{document}